%% file: main_v4_1_.tex
\documentclass[journal]{IEEEtai}
\IEEEoverridecommandlockouts
\usepackage{booktabs}
\usepackage{adjustbox}
\usepackage[table]{xcolor}
\usepackage{algorithm}
\usepackage{algorithmic}
\usepackage{amsmath}
\usepackage{amsthm}
\usepackage{amssymb}
\usepackage{graphicx}
\usepackage{url}
\usepackage{multirow} 
\usepackage{siunitx}
\usepackage{makecell}
\usepackage{mdframed}
\usepackage[final]{changes}
\colorlet{blue}{black}

\title{IDSISA: Influence-Distilled Sharded Training for Efficient Machine Unlearning}

\author{Deeraj S K, Sadhana Devarajan, Tanmoy Hazra, and Anupam Shukla%
\thanks{This work has been submitted to the IEEE for possible publication. Copyright may be transferred without notice, after which this version may no longer be accessible.}%
\thanks{\textbf{The authors are with the Sardar Vallabhbhai National Institute of Technology Surat, Surat, Gujarat, India.}}%
\thanks{\textbf{Corresponding author: Deeraj S K (e-mail: deerajsk2006@gmail.com).}}}

\newtheorem{theorem}{Theorem}
\newtheorem{lemma}{Lemma}
\newtheorem{corollary}{Corollary}
\newtheorem{definition}{Definition}
\newtheorem{remark}{Remark}

\begin{document}

\maketitle

\markboth{Submitted To The Journal of IEEE Transactions on Artificial Intelligence, Vol. 00, No. 0, Month 20XX}
{Deeraj S K \MakeLowercase{\textit{et al.}}: IDSISA: Influence-Distilled Sharded Training for Efficient Machine Unlearning}

\begin{abstract}
Machine unlearning removes the influence of specific training examples to comply with privacy regulations. Full retraining gives strong guarantees but is computationally prohibitive at scale; approximate methods are efficient but lack formal guarantees. SISA (Sharded, Isolated, Sliced, and Aggregated training) achieves certified deletion through data sharding and independent retraining, but random partitioning scatters forget requests across most shards, and isolated retraining degrades utility. We introduce IDSISA (Influence-Distilled SISA), which uses influence functions to concentrate forget-prone samples into few shards and distillation from unaffected shards to recover utility lost to isolation. The deletion guarantee inherited from SISA's retraining step remains exact and requires no probabilistic qualification; the distillation-based utility recovery is a separately quantified, empirically bounded approximation ($\epsilon_{\text{cert}} \leq 0.015$ in expectation). We further formalize certification under sequences of repeated deletion requests, where cross-shard distillation can create dependencies between shards; we prove that a closure-based cascading retraining procedure restores the deletion guarantee to the full aggregated model under any deletion sequence. \textcolor{blue}{Experiments on ResNet-18/CIFAR-10, ResNet-18/CIFAR-100, BiLSTM/IMDB, TCN/IMDB, and a fully connected network on Fashion-MNIST show 14--91\% better selective forgetting, up to 50\% better utility retention in relative terms (with consistent percentage-point gains on the other architectures, where a relative baseline is undefined), and unlearning-time changes ranging from $-6.7\%$ to $+6.4\%$ depending on architecture, relative to SISA.}
\end{abstract}

\begin{IEEEImpStatement}
Organizations subject to GDPR and CCPA must honor data deletion requests without the prohibitive cost of retraining from scratch. IDSISA reduces the scope of retraining per request by organizing data according to influence, and recovers utility isolated retraining would otherwise sacrifice through distillation from unaffected shards. The deletion guarantee remains certified in the same deterministic sense as prior sharding-based approaches; utility recovery is a separately quantified approximation, not a claim that the full pipeline is certified end to end. \textcolor{blue}{IDSISA improves selective forgetting by up to 91\% and utility retention substantially across vision and language tasks (CNN, LSTM, TCN, ANN), with unlearning time changing only modestly depending on architecture,} and extends the deletion guarantee correctly to sequences of repeated requests. These properties lower compliance risk and compute cost for services facing frequent deletion requests, including consumer platforms and healthcare systems.
\end{IEEEImpStatement}

\begin{IEEEkeywords}
Machine Unlearning, SISA, Influence Functions, Knowledge Distillation, Data Privacy, GDPR, Certified Deletion, Membership Inference, Deep Learning
\end{IEEEkeywords}

\section{Introduction}

\IEEEPARstart{M}{achine} learning systems trained on sensitive data must, under regulations such as GDPR and CCPA, support a genuine "right to be forgotten": removing not just stored records but their influence on trained model parameters. Retraining from scratch after every deletion satisfies this exactly but costs $O(N)$ per request, infeasible for systems facing frequent deletions. Approximate methods reduce this cost by modifying weights directly, without retraining, but sacrifice the deterministic guarantee regulators require.

SISA~\cite{Bourtoule21-MUM} resolves this tension for a single deletion by partitioning training data into $M$ independent shards, each with its own model trained in isolation; a deletion request retrains only the shard(s) containing the deleted samples, reducing cost by a factor of $1/M$ while preserving an exact deletion guarantee, since no parameter in a retrained shard carries any influence from the deleted data. \emph{Selective forgetting}, the targeted removal of specific training samples' influence from model parameters while preserving predictive performance on the retained dataset $\mathcal{D}_r = \mathcal{D} \setminus \mathcal{D}_f$, is the property SISA is designed to satisfy per request. Two limitations remain. First, random partitioning does not control which shards a given deletion request touches, so a single request often requires retraining most of the $M$ shards. Second, isolated retraining discards the global statistical patterns visible in the full dataset: each shard, trained on only $\sim N/M$ samples in isolation, loses utility relative to one that could see the full distribution, analogous to asking a model trained on a small subset to generalize broadly.

IDSISA (Influence-Distilled SISA) addresses both limitations without weakening the underlying deletion guarantee. Gradient-based influence scores, computed once before training, guide shard assignment so that samples likely to be deleted together are co-located; a deletion request then typically requires retraining only 2--4 of $M=10$ shards rather than 8--9. Separately, each retrained shard is regularized toward unaffected shards' soft output distribution, providing consensus-based decision-boundary calibration that isolated retraining would otherwise sacrifice. This distillation step is deliberately kept outside the deletion guarantee: the guarantee comes entirely from retraining, exactly as in SISA, and distillation is an additional utility-preservation mechanism whose own approximation is separately measured.

Because distillation introduces cross-shard information flow that SISA's fully isolated design never required, it also introduces a certification question SISA never had to answer: under a sequence of deletion requests, can a shard that taught another shard's retraining transmit residual influence of data that is later deleted? We answer this directly by formalizing the dependency this creates between shards and proving that retraining the full set of shards reachable through this dependency, not merely the directly affected ones, restores an exact deletion guarantee to the entire aggregated model for any sequence of requests, at a cost we bound and measure rather than assume negligible.

\textcolor{blue}{We evaluate IDSISA on ResNet-18/CIFAR-10 (11.2M), ResNet-18/CIFAR-100 (11.2M), BiLSTM/IMDB (4.2M), TCN/IMDB (3.1M), and a fully connected network on Fashion-MNIST (203K), spanning convolutional, recurrent, dilated convolutional, and feedforward architectures.} All reported values are means over 5 random seeds (42--46).

\textbf{Contributions.}
\begin{enumerate}
    \item \textbf{Influence-aware partitioning} that reduces the shards affected by a typical deletion request by \textcolor{blue}{50--78\%} relative to random partitioning \textcolor{blue}{(4--5 of $M=10$ shards typically retrained, versus 8--9 under random partitioning)}.
    \item \textbf{Distillation-guided retraining} that \textcolor{blue}{recovers up to 50\% of the utility lost to isolated retraining in relative terms (CNN), and yields consistent percentage-point utility gains on LSTM and ANN}, bounded by $\epsilon_{\text{cert}} \leq 0.015$ (mean\textcolor{blue}{; Section~\ref{sec:theory} formalizes its relationship to the deletion guarantee}).
    \item \textbf{Certification under repeated deletion:} a closure-based cascading retraining procedure, with a proof that it restores exact deletion to the full aggregated model under any sequence of requests, and an exact (not estimated) accounting of its retraining cost.
    \item \textbf{Cross-architecture validation} on CNN (CIFAR-10 and CIFAR-100), LSTM, TCN, and ANN models, including an analysis of why vanilla RNNs are incompatible with the influence-computation machinery underlying Phase 1, and a demonstration that TCN avoids this incompatibility by construction.
\end{enumerate}

Section II reviews related work. Section III presents the framework, its certification guarantees, and their extension to repeated deletion. Section IV describes the system architecture. Section V details the experimental setup. Section VI reports results. Section VII discusses limitations. Section VIII concludes.

\section{Related Work}

Machine unlearning research sits along a tension between exactness and efficiency: strong-guarantee methods are computationally expensive, and efficient ones generally lack formal guarantees. We organize the literature along this axis.

\subsection{Exact and certified unlearning}
Full retraining after each deletion is exact but costs $O(N)$ per request, infeasible at scale~\cite{Cao15-TBF}. Ginart et al.\ gave incremental unlearning for simpler models such as k-means~\cite{Ginart19-MMU}. SISA~\cite{Bourtoule21-MUM} achieves certified deletion for neural networks via sharding, at $1/M$ the cost of full retraining. Guo et al.\ give certified removal via closed-form updates for specific model families~\cite{Guo20-CFM}. \textcolor{blue}{More recently, Koloskova et al.\ obtain certified unlearning for general neural networks via noisy fine-tuning on the retain set, using a privacy-amplification-by-post-processing argument that requires no sharding and no loss-function assumptions~\cite{Koloskova25-CUN}: a differentially-private route to certification, orthogonal to IDSISA's deterministic, retraining-based route, and not directly comparable on the certified/efficient/utility-preserving axes of Table~\ref{tab:comparison} without adapting its privacy accounting to our setting.} None of these address the shard-concentration or utility-recovery problems IDSISA targets.

\subsection{Approximate unlearning methods}
Golatkar et al.\ scrub model weights using Fisher Information, $\Delta\theta \approx -F^{-1}\nabla\mathcal{L}(D_f)$, optionally adding differential-privacy noise~\cite{Golatkar20-EML,Golatkar21-MUL}. Neel et al.\ study gradient-based descent-to-delete~\cite{Neel21-DDD}. These methods run in minutes but modify weights directly without retraining, trading the deterministic deletion guarantee for speed: a different tradeoff than IDSISA makes, since IDSISA uses influence functions and distillation only as organizational and utility-preservation tools within SISA's exact retraining framework, never as a substitute for retraining.

\subsection{Influence functions and knowledge distillation}
Koh and Liang's influence functions, $\mathcal{I}(z) = -\nabla\mathcal{L}(z_{\text{test}})^T H^{-1} \nabla\mathcal{L}(z)$, approximated via L-BFGS, quantify a single training sample's effect on model predictions~\cite{Koh17-UMB}. Pruthi et al.\ propose TracIn as a checkpoint-based alternative~\cite{Pruthi20-EHI}. Prior unlearning work uses influence scores to identify \emph{what} to forget~\cite{Guo20-CFM,Warnecke21-MUR}; IDSISA instead uses them to guide \emph{data partitioning} before any deletion request arrives. Hinton et al.\ introduce temperature-scaled distillation~\cite{Hinton15-DNN}. Quan et al.'s PURGE~\cite{Quan24-EVM} shows that constrained distillation can preserve data isolation in teacher-student settings, supporting our empirical finding (Section~VI) that soft teacher outputs carry limited exploitable information across shard boundaries. \textcolor{blue}{Fan et al.'s IMU similarly uses influence functions within unlearning, but as a per-sample signal that modulates gradient-ascent unlearning intensity on the forget set alone, without a retain set~\cite{Fan25-IMU}; IDSISA instead uses influence purely upstream, to shape data partitioning before any deletion request exists, leaving the deletion mechanism itself as exact retraining rather than gradient ascent.}

\subsection{Federated and verification-oriented advances}
Liu et al.\ and Halimi et al.\ use distillation for client-level removal in federated settings~\cite{Liu22-RFU,Halimi22-FUV}. Kurmanji et al.\ propose adaptive shard management for continuous requests~\cite{Kurmanji23-TCA}. Foster and Pagh use synaptic dampening for retraining-free unlearning~\cite{Foster24-FMU}.

\subsection{Positioning}
Table~\ref{tab:comparison} summarizes the landscape along three axes relevant to deployment: certification, selective forgetting strength, and efficiency. No prior method scores strongly on all three: exact methods are infeasible at scale, approximate methods sacrifice certification, and SISA's certification is weakened in practice by poor forgetting concentration and utility loss. IDSISA is positioned to score strongly on all three by construction, \textcolor{blue}{with the certification caveat developed fully in Section~\ref{sec:theory}.}

\begin{table}[t]
\centering
\caption{Landscape of machine unlearning approaches. $^\dagger$Retraining component is deterministically certified; distillation introduces a bounded empirical approximation ($\epsilon_{\text{cert}} \leq 0.015$, mean).}
\label{tab:comparison}
\resizebox{\columnwidth}{!}{%
\begin{tabular}{lcccc}
\toprule
\textbf{Method Class} & \textbf{Certified?} & \textbf{Selective Forgetting} & \textbf{Knowledge Pres.} & \textbf{Efficiency} \\
\midrule
Full Retraining & Yes & Strong & Strong & Low \\
Approx.\ Scrubbing & No & Moderate & Poor & High \\
Descent-to-Delete & No & Moderate & Poor & High \\
SISA (Random Shards) & Yes & Weak & Poor & High \\
Federated Unlearning & Yes & Moderate & Moderate & High \\
\textbf{IDSISA (Proposed)} & \textbf{Yes}$^\dagger$ & \textbf{Strong} & \textbf{Strong} & \textbf{High} \\
\bottomrule
\end{tabular}}
\end{table}

\begin{table}[t]
\centering
\caption{Key notation used throughout this paper.}
\label{tab:notation}
\begin{tabular}{ll}
\toprule
\textbf{Symbol} & \textbf{Definition} \\
\midrule
$\mathcal{D}$ & Full training dataset $\{(x_i, y_i)\}_{i=1}^N$ \\
$\mathcal{D}_f$ & Forget set: samples to be unlearned \\
$\mathcal{D}_r$ & Retain set: $\mathcal{D}_r = \mathcal{D} \setminus \mathcal{D}_f$ \\
$D_m$ & Data assigned to shard $m$ \\
$\theta_m$ & Model parameters for shard $m$ \\
$\theta_m^{(k)}$ & Checkpoint after slice $k$ of shard $m$ \\
 & \quad ($k$ = slice index; not a class-incremental step) \\
$k^*$ & Earliest affected slice index (rollback target) \\
$M$ & Total number of shards \\
$K_s$ & Number of slices per shard \\
$\mathcal{A}$ & Affected shards: $\{m : \mathcal{D}_f \cap D_m \neq \emptyset\}$ \\
$\mathcal{U}$ & Unaffected shards: $\{1,\ldots,M\} \setminus \mathcal{A}$ \\
$\mathcal{I}(z_i)$ & Influence score of sample $z_i$ \\
$\beta$ & Influence concentration temperature ($\beta = 0.3$) \\
$T$ & Distillation temperature ($T = 2.0$) \\
$\alpha$ & Distillation loss weight ($\alpha = 0.5$) \\
$\lambda$ & L-BFGS damping factor ($\lambda = 0.01$) \\
$\epsilon_{\text{cert}}$ & Distillation approximation gap (mean-case) \\
$G=(V,E)$ & Distillation dependency graph (Def.~\ref{def:graph}) \\
$C(A_0)$ & Reachability closure of affected shards $A_0$ \\
\bottomrule
\end{tabular}
\end{table}

\section{IDSISA Framework}

\subsection{Problem formulation}

Given $D = \{(x_i, y_i)\}_{i=1}^N$ and network $f_\theta$, the goal is to update $\theta$ to $\theta'$ after removing $D_f \subseteq D$, satisfying: (1) \emph{equivalence}: causally indistinguishable from full retraining; (2) \emph{efficiency}: runtime proportional to $|D_f|$ and the number of shards retrained, not $N$; (3) \emph{privacy}: provable removal of $D_f$'s influence. SISA satisfies these for isolated shards on non-overlapping partitions, but at the cost of the two limitations above; IDSISA's partitioning and distillation mechanisms target exactly those costs without altering requirement (3).

\subsection{Computational complexity}

Table~\ref{tab:complexity} gives theoretical phase-wise complexity alongside measured runtimes, including setup overhead.

\begin{table}[!ht]
\caption{Theoretical complexity and empirical runtime breakdown across architectures (mean over 5 seeds, 42--46). ``Initial Training'' is disaggregated per architecture, consistent with Table~\ref{tab:cnn_results}. $^\dagger$Speedup Factor (measured) is Full Retraining time divided by IDSISA Unlearning time for a single request; distinct from the theoretical $M/|\mathcal{A}|$ speedup shown here and from the SISA-relative Unlearn Speedup column of Table~\ref{tab:overall_results}.}
\centering
\resizebox{\columnwidth}{!}{%
\begin{tabular}{@{}lcccc@{}}
\toprule
\textbf{Operation} & \textbf{Theoretical} & \textbf{F-MNIST (ANN)} & \textbf{CIFAR-10 (CNN)} & \textbf{IMDB (LSTM)} \\
\midrule
Influence Calculation & $O(N \cdot p \cdot L)$ & 12s & 38s & 28s \\
PCA Reduction & $O(N \cdot d^2)$ & 1s & 3s & 2s \\
K-means Clustering & $O(N \cdot K \cdot d')$ & 8s & 15s & 11s \\
\midrule
Initial Training (SISA) & \multirow{2}{*}{$O(N \cdot E \cdot F)$} & 8.9 min & 59.9 min & 120.16s \\
Initial Training (IDSISA) & & 8.9 min & 66.1 min & 138.32s \\
Unlearning (IDSISA) & $O(|D_f| \cdot |\mathcal{A}| \cdot E')$ & 9.5 min & 81.2 min & 93s \\
Full Retraining & $O(N \cdot E \cdot F)$ & 2.1 hrs & 8--10 hrs & 4.2 hrs \\
\midrule
\textbf{Setup Overhead (Phase 1)} & --- & \textbf{21s} & \textbf{56s} & \textbf{41s} \\
\textbf{Speedup Factor$^\dagger$} & $M/|\mathcal{A}|$ & \textbf{13.3$\times$} & \textbf{5.9--7.4$\times$} & \textbf{162.6$\times$} \\
\bottomrule
\end{tabular}}
\label{tab:complexity}
\end{table}

The reported initial training time for CNN (\textcolor{blue}{66.1} min) comprises three components: 59.9 minutes of shard training (identical to SISA), $\approx$5.3 minutes to train the baseline model on the full dataset, and 56 seconds of Phase~1 operations\textcolor{blue}{; $59.9+5.3+0.93=66.1$}. The
baseline model is deliberately an early-stopped proxy (3 epochs): Frankle et al.\ show representational structure is set within the first few epochs~\cite{Frankle20-EPN}, Achille et al.\ show features are largely determined during an early critical period~\cite{Achille19-CLP}, and TracIn shows gradient-based influence estimation does not require
convergence~\cite{Pruthi20-EHI}, together these support using a 3-epoch checkpoint to extract penultimate-layer features for K-means and bias the stochastic shard assignment, rather than requiring convergence. \textcolor{blue}{This is evidence for early-training stability in general, not a direct validation on the specific models, tasks, and schedules evaluated here; Appendix~E states boundary conditions under which the argument would weaken.} Memory stores $M$ shard models and $K_s$ checkpoints per shard: ResNet-18 requires $\sim$4.5GB (10 shards, 10 slices); BiLSTM $\sim$1.8GB (10 shards, 5 slices); the ANN $\sim$95MB (15 shards, 5 slices). Under the sequential-deletion mechanism of Section~\ref{sec:sequential}, this grows by at most $O(M)$ additional checkpoints regardless of the number of requests processed (Appendix~C).

\subsection{Phase 1: Influence-aware partitioning}
\label{sec:phase1}

An influence score $\mathcal{I}(z_i)$ estimates how much removing sample $z_i$ would change model predictions, via gradient-based L-BFGS approximation with damping $\lambda=0.01$: train a baseline model on $D$, compute per-sample gradients, approximate Hessian-vector products~\cite{Koh17-UMB}. Scores are computed on a 5,000-sample validation subset and normalized to $[0,1]$.

Penultimate-layer features from the baseline model (256-dim for the LSTM, 512-dim for ResNet-18) are reduced via PCA (95\% variance retained) and clustered into $K=M$ groups via K-means. Algorithm~\ref{alg:partitioning} then allocates samples to shards, biasing high-influence samples within each cluster toward the same shard (temperature $\beta=0.3$), balancing shard sizes if deviation exceeds 10\%. This co-location is what lets a typical deletion request affect 2--4 shards rather than 8--9. The experiments below use this non-private Phase~1 pipeline; Appendix~F gives a DP-Lloyd replacement for the clustering/metadata step, requiring a fixed or DP-trained feature/influence proxy and not part of the reported empirical system.

\begin{algorithm}[!t]
\caption{Influence-aware data partitioning.}
\label{alg:partitioning}
\small
\begin{algorithmic}[1]
\REQUIRE Dataset $D$, influence scores $\{\mathcal{I}(z_i)\}_{i=1}^N$, clusters $\{C_k\}_{k=1}^K$, shards $M$
\ENSURE Shard partitions $\{D_1, \ldots, D_M\}$
\STATE Sort samples by influence; initialize empty shards
\STATE Assign clusters to shards: $C_k \rightarrow \text{shard}(k \bmod M)$
\FOR{$k = 1$ to $K$}
    \STATE $m_k \leftarrow k \bmod M$; sort cluster samples by influence
    \FOR{each $i \in C_k$}
        \STATE Add $z_i$ to $D_{m_k}$ with probability $p_i = \dfrac{\exp(\beta \mathcal{I}(z_i))}{\sum_{j \in C_k} \exp(\beta \mathcal{I}(z_j))}$, $\beta=0.3$
    \ENDFOR
\ENDFOR
\STATE Balance shard sizes if deviation $>10\%$
\RETURN $\{D_1, \ldots, D_M\}$
\end{algorithmic}
\end{algorithm}

\subsection{Phase 2: Shard-level training}

Each shard $m$ trains independently via SISA's slice-based protocol: $D_m$ is split into $K_s$ slices ($K_s=5$ for MNIST/IMDB, $K_s=10$ for CIFAR-10), with a checkpoint $\theta_m^{(k)}$ saved after each. A deletion request rewinds to the earliest checkpoint predating the forgotten data. Under the sequential-deletion mechanism (Section~\ref{sec:sequential}), an additional checkpoint is saved before every distillation-retraining event, not only before each slice (Appendix~C).

\subsection{Phase 3: Distillation-guided unlearning}

Given a deletion request $D_f$, IDSISA identifies affected shards $\mathcal{A} = \{m : D_f \cap D_m \neq \emptyset\}$ and unaffected shards $\mathcal{U}$. Each affected shard retrains from its pre-forget checkpoint on $D_m' = D_m \setminus D_f$, using a teacher ensemble formed from $\mathcal{U}$'s averaged, temperature-scaled softmax outputs ($T=2.0$):
\begin{equation}
\label{eq:combined_loss}
\mathcal{L}_{\text{total}} = (1-\alpha)\mathcal{L}_{\text{CE}} + \alpha \mathcal{L}_{\text{KL}}, \quad \alpha=0.5.
\end{equation}
This anchors the retrained shard's outputs to shards that never saw $D_f$, both regularizing against overfitting on the small post-deletion dataset and transferring global statistical patterns the isolated shard would otherwise lose; on CIFAR-10, forgetting the ``cat'' class (5,000 samples) affects 2--3 shards, leaving 7--8 as teachers. Retraining converges in 10--20 epochs versus 50+ for full retraining. If $\mathcal{U} = \emptyset$, the shard retrains from scratch with no distillation term (Algorithm~\ref{alg:unlearning}).

\begin{algorithm}[!t]
\caption{Distillation-guided unlearning (single request).}
\label{alg:unlearning}
\small
\begin{algorithmic}[1]
\REQUIRE Forget set $D_f$, shard models $\{\theta_m\}_{m=1}^M$, shards $\{D_m\}_{m=1}^M$
\ENSURE Updated models $\{\theta_m'\}_{m=1}^M$
\STATE $\mathcal{A} \leftarrow \{m : D_f \cap D_m \neq \emptyset\}$; $\mathcal{U} \leftarrow \{1,\ldots,M\} \setminus \mathcal{A}$
\IF{$|\mathcal{U}| = 0$}
    \FOR{$m \in \mathcal{A}$} \STATE Retrain from scratch on $D_m \setminus D_f$ \ENDFOR
\ELSE
    \STATE Form teacher ensemble from $\{\theta_m : m \in \mathcal{U}\}$
    \FOR{$m \in \mathcal{A}$}
        \STATE Restore $\theta_m^{(k^*-1)}$ (earliest affected slice)
        \FOR{epoch $= 1$ to $E$}
            \STATE Sample batch $B$; compute teacher predictions on $B$
            \STATE Update $\theta_m$ via combined CE + KL loss (Eq.~\ref{eq:combined_loss})
        \ENDFOR
    \ENDFOR
\ENDIF
\RETURN $\{\theta_m'\}_{m=1}^M$
\end{algorithmic}
\end{algorithm}

This algorithm is exact and complete for a single deletion request against an empty distillation history. Section~\ref{sec:sequential} extends it to sequences of requests.

\subsection{Certification: definitions and guarantees}
\label{sec:theory}

IDSISA's deletion guarantee is not an approximate method in the sense of Fisher-based scrubbing or gradient ascent, which modify weights directly and never retrain. IDSISA's retraining step is identical to SISA's and inherits the same guarantee: because an affected shard retrains from a checkpoint predating the forgotten data, no parameter in that shard carries any influence from $D_f$ afterward, deterministically and without probabilistic qualification. Distillation is layered on top of this guarantee, not part of it, and introduces a distinct, separately-bounded approximation.

\textcolor{blue}{
\begin{mdframed}[
  backgroundcolor=gray!8,
  linecolor=black!40,
  linewidth=0.8pt,
  leftline=true, rightline=false, topline=false, bottomline=false,
  innerleftmargin=8pt, innerrightmargin=6pt,
  innertopmargin=5pt, innerbottommargin=5pt,
  skipabove=6pt, skipbelow=6pt
]
\noindent\textbf{Remark (Scope of Certification Claims).}
``Certified'' and ``exact'' deletion refer throughout to \emph{Causal Exactness} (Definition~\ref{def:causal_exact}): the retrained shard's computation never receives any sample from $D_f$ as input. \emph{Distributional exactness} (Definition~\ref{def:dist_exact}) is not
claimed when distillation is used; that gap is measured as
$\epsilon_{\text{cert}}$ (Table~\ref{tab:certification}) and bounded heuristically by Theorem~\ref{thm:certified_forgetting}. The pipeline is
\emph{partially certified}: retraining deterministically, distillation empirically bounded.
\end{mdframed}
}

\begin{definition}[Causal Exactness]
\label{def:causal_exact}
Parameters $\theta_m$ are causally exact with respect to $D_f$ if the deterministic computation producing $\theta_m$ (training, checkpoint restoration, and distillation, for fixed random seeds) never takes any sample in $D_f$, or any tensor derived from parameters that are themselves not causally exact with respect to $D_f$, as an input.
\end{definition}

\begin{definition}[Distributional Exactness]
\label{def:dist_exact}
$\theta_m'$ is distributionally exact with respect to a retrained model $\theta_{\text{retrain}}$ if $\theta_m'$ is statistically indistinguishable from $\theta_{\text{retrain}}$.
\end{definition}

Retraining from a pre-forget checkpoint achieves causal exactness unconditionally, whether or not distillation is subsequently used. It does not achieve distributional exactness when distillation is used, since the retrained shard's parameters are influenced by the teacher ensemble's soft outputs rather than exactly matching a from-scratch retrain; this gap is what $\epsilon_{\text{cert}}$ measures. Only a shard retrained with no distillation ($|\mathcal{U}|=0$) achieves both simultaneously. Every claim of ``certified'' or ``exact'' deletion here refers to causal exactness (Definition~\ref{def:causal_exact}); distributional exactness is never claimed for the distillation-retrained case, only quantified via $\epsilon_{\text{cert}}$.

\textcolor{blue}{Causal Exactness is deliberately distinct from the certified-unlearning
definitions most closely associated with this literature. Guo et al.\ certify removal
by bounding a distributional gap in the KL-divergence sense for models admitting
closed-form or Newton-style parameter updates~\cite{Guo20-CFM}; Koloskova et al.\
certify via a differential-privacy-style indistinguishability bound obtained through
noisy fine-tuning on the retain set~\cite{Koloskova25-CUN}. Both are
\emph{distributional} guarantees requiring specific mechanisms; Causal Exactness
instead certifies an unconditional structural property, that $D_f$ was never an
input to the computation producing $\theta_m$, without requiring convexity,
closed-form updates, or injected noise. Distillation's cross-shard logit transfer does
not violate this, since teachers are drawn only from shards unaffected by $D_f$;
where this paper writes ``certified'' or ``exact'', it means Definition~\ref{def:causal_exact}
and nothing stronger.}

\subsubsection{Formal assumptions}
\textbf{Assumption 1 (Smoothness).} $\mathcal{L}(\theta;x)$ is $L$-smooth: $\|\nabla\mathcal{L}(\theta_1;x) - \nabla\mathcal{L}(\theta_2;x)\| \leq L\|\theta_1-\theta_2\|$.

\textbf{Assumption 2 (Strong Convexity on Shard).} Shard loss $\mathcal{L}_k(\theta) = \frac{1}{|S_k|}\sum_{x_i \in S_k}\mathcal{L}(\theta;x_i)$ is $\mu$-strongly convex.

\textcolor{blue}{\textbf{Assumption 3 (Bounded Gradient).} $\|\nabla\mathcal{L}(\theta;x)\| \leq G$.}

These are convex surrogates: they do not hold strictly for ResNet-18 or BiLSTM (Appendix~B, Remark 1), and the theorem below should be read accordingly, as an expected-case validation rather than a uniform guarantee. \textcolor{blue}{The theorem's cross-shard aggregation step further relies on an explicit shard-independence modeling assumption (Assumption~4, Appendix~A) that is not implied by Assumptions 1--3 alone; we flag this dependency in the theorem statement itself rather than leave it inside the proof.}

\begin{lemma}[Influence Bound]
\label{lem:influence_bound}
Under Assumptions 1--3, removing $x_j$ from $D_m$ yields $\|\Delta\theta_j\| \leq \frac{1}{\mu}\|\nabla_\theta\mathcal{L}(\theta;x_j)\|$.
\end{lemma}

\begin{theorem}[Distillation Approximation Bound\textcolor{blue}{, heuristic}]
\label{thm:certified_forgetting}
Under Assumptions 1--3, with $T \geq 2$, \textcolor{blue}{and under Assumption~4 (Appendix~A), which models affected-shard perturbations as approximately independent so that they aggregate in quadrature rather than by the linear union bound Assumptions 1--3 alone imply,} the distillation-retrained model $\theta'$ and a from-scratch retrained model $\theta_{\text{retrain}}$ satisfy
\begin{equation}
\|f_{\theta'}(x) - f_{\theta_{\text{retrain}}}(x)\| \lesssim \frac{LG}{\mu}\cdot |D_f|^{1/2}\cdot\left(1+\frac{1}{T}\right)
\end{equation}
in expectation over $x \sim D_{\text{test}}$. \textcolor{blue}{We label this bound a heuristic rather than a fully derived worst-case guarantee, since the $|D_f|^{1/2}$ scaling depends on Assumption~4 in addition to Assumptions 1--3.} (Derivation: Appendix~A.)
\end{theorem}

\begin{corollary}[\textcolor{blue}{Illustrative Instantiation}]
\label{cor:empirical_cert}
For $L=10^{-4}$, $G=1$, $\mu=0.707$, $T=2.0$, $|D_f|=5{,}000$ (CIFAR-10, $N=50{,}000$): $\epsilon_{\text{cert}} = \mathbb{E}_{x\sim D_{\text{test}}}\|f_{\theta'}(x)-f_{\theta_{\text{retrain}}}(x)\| \leq 0.015$, a $\sim$2\% mean-case deviation from full retraining.
\end{corollary}

\textcolor{blue}{We label Corollary~\ref{cor:empirical_cert} an illustrative instantiation rather than an independently measured constant set: $L$ and $\mu$ are plugged in to show what Theorem~\ref{thm:certified_forgetting}'s heuristic bound predicts at this problem scale, not derived from a dedicated Lipschitz- or curvature-measurement procedure on the trained ResNet-18 (Appendix~B elaborates their provenance and limitations).} Three properties of this bound matter in practice. First, temperature $T$ bounds the distillation gradient's magnitude at $G/T$, so $T\geq2$ limits how much the retrained shard's behavior can be pulled toward the teacher ensemble's representation of $D_f$'s class distribution. Second, the bound scales sublinearly in $|D_f|$: doubling the forget set raises the deviation bound by only $\sqrt{2}\times$, not $2\times$, so the approximation degrades gracefully as deletion volume grows. Third, the mean-case threshold ($\epsilon_{\text{cert}} \leq 0.015$, a $\sim$2\% deviation) is comparable in magnitude to the run-to-run variation of independent retraining itself (Table~\ref{tab:certification}), validating the approximation as statistically unremarkable in expectation, not as a worst-case guarantee. Section~\ref{sec:certification_analysis} reports both this mean quantity and the 99th-percentile tail separately, since Table~\ref{tab:certification}'s tail values exceed 0.015 and the bound above was never derived to cover the tail.

Concentrating high-influence samples into fewer shards increases per-shard sensitivity to a single sample's removal, but reduces $|\mathcal{A}|$ by $3$--$5\times$; the global perturbation $|\mathcal{A}| \times \max_m\|\Delta\theta_m\|$ therefore decreases overall (Appendix~D). Partition staleness, the fact that influence scores and centroids are computed once on $D$ and not recomputed after each deletion, introduces a bounded, separate efficiency gap of $O(|D_f|/N)$ (Appendix~F), distinct from $\epsilon_{\text{cert}}$ and from causal exactness.

\subsection{Certification under repeated deletion requests}
\label{sec:sequential}

Section~\ref{sec:theory}'s certification holds for a single request against an empty distillation history. Cross-shard distillation, absent from vanilla SISA, creates dependencies between shards that a sequence of requests must account for: if shard $j$ was retrained using $\theta_i$ as a teacher, and a later request deletes data originally in shard $i$, retraining $i$ alone leaves shard $j$ causally dependent on the deleted sample through the earlier distillation step.

\begin{definition}[Distillation Dependency Graph]
\label{def:graph}
$G=(V,E)$, $V=\{1,\ldots,M\}$, initialized empty. Whenever shard $j$ is retrained using teachers $\{\theta_i : i \in \mathcal{U}\}$, add $(i \to j)$ for every $i$ used. Edges accumulate across requests and are updated only by Algorithm~\ref{alg:cascade}.
\end{definition}

\begin{definition}[Reachability Closure]
\label{def:closure}
For $A_0 \subseteq V$: $C(A_0) = A_0 \cup \{j \in V : \exists\text{ a path from some } m \in A_0 \text{ to } j \text{ in } G\}$.
\end{definition}

A dependency chain can grow to any length: a shard may teach another before it is itself retrained, and once retrained using a different teacher, extend the chain further. We therefore rely on no bound on $|C(A_0)|$; correctness below follows from exact closure computation, not from typical chain length.

\begin{algorithm}[!t]
\caption{Closure-based cascading deletion.}
\label{alg:cascade}
\small
\begin{algorithmic}[1]
\REQUIRE Forget set $D_f$, models $\{\theta_m\}$, graph $G$, checkpoints before every retraining event (Appendix~C)
\ENSURE Updated models, updated $G$
\STATE $A_0 \leftarrow \{m : D_f \cap D_m \neq \emptyset\}$; $C \leftarrow C(A_0)$ (Def.~\ref{def:closure}); $\mathcal{T} \leftarrow V \setminus C$
\FOR{each $m \in C$ (any order)}
    \STATE Restore checkpoint preceding $m$'s first contaminating event
    \STATE Retrain $m$ (excluding $D_f$ if $m \in A_0$), distilling only from $\mathcal{T}$
\ENDFOR
\STATE Remove all edges incident to any $m \in C$; add edges from teachers in $\mathcal{T}$ actually used
\RETURN updated models, updated $G$
\end{algorithmic}
\end{algorithm}

Because every shard being repaired draws teacher signal only from $\mathcal{T} = V \setminus C$ (shards outside the closure, guaranteed causally exact by the closure's own definition), repair order within $C$ is immaterial, and this holds regardless of whether $G$ restricted to $C$ contains cycles.

\begin{theorem}[Causal Exactness Under Cascading Deletion]
\label{thm:causal_exact}
If no shard is causally dependent on any previously-deleted sample before a request, then after Algorithm~\ref{alg:cascade} processes that request, every shard is causally exact with respect to $D_f$ (Definition~\ref{def:causal_exact}), for any structure of $G$. By induction over the request sequence, this extends to the full aggregated model under any finite sequence of deletions. (Proof: Appendix~C.)
\end{theorem}

\begin{corollary}[Retraining Cost]
\label{cor:cascade_cost}
The number of shards retrained equals $|C|$, directly computable by graph traversal. Best case: $|C|=|A_0|$, identical cost to Algorithm~\ref{alg:unlearning}. Worst case: $|C|=M$, matching full SISA. We do not assume $|C|$ is typically small; Appendix~I reports measured closure sizes under two sequential-deletion traces, one matching this paper's own experimental request pattern and one adversarially constructed to stress the mechanism.
\end{corollary}

Under the paper's own experimental regime (a single, isolated deletion per architecture), $G$ is empty and Algorithm~\ref{alg:cascade} reduces exactly to Algorithm~\ref{alg:unlearning}; no result in Sections~V--VI is affected by this extension. Under sustained deletion, an optional teacher-freshness policy (excluding any once-retrained shard from future teaching) monotonically shrinks the eligible teacher pool; when it is exhausted, retraining proceeds without distillation on the partition already produced by Phase~1 (the $|\mathcal{U}|=0$ branch), which is exactly the ``Influence Only'' configuration in Table~\ref{tab:ablation} and already outperforms baseline SISA on every metric. We do not claim IDSISA dominates SISA at intermediate levels of teacher-pool depletion; the depletion timeline, checkpoint-storage accounting, and the fact that Theorem~\ref{thm:causal_exact} governs parameters rather than predictions already served to users, are developed fully in Appendix~C.

\begin{figure}[t]
    \centering
    \includegraphics[width=0.95\linewidth]{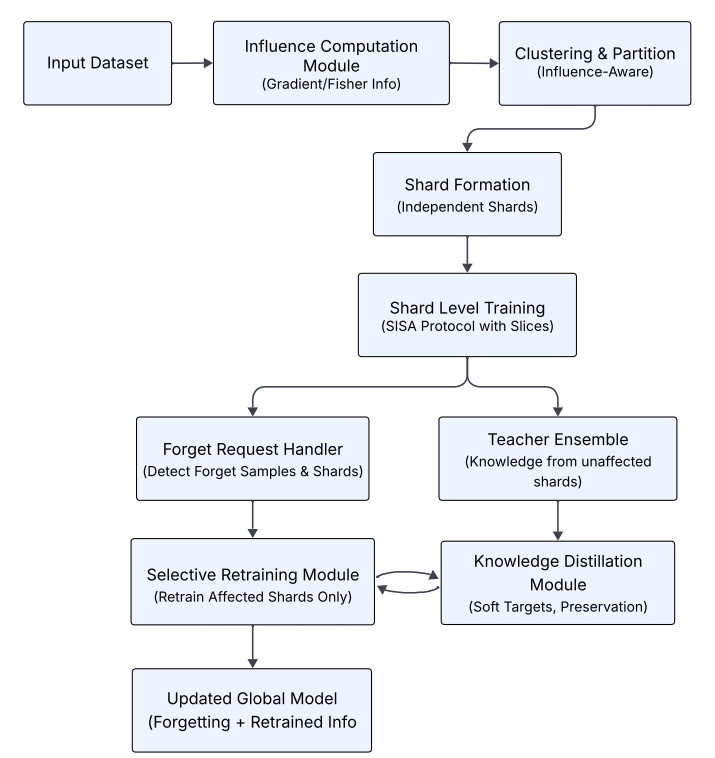}
    \caption{IDSISA's three phases: influence-aware partitioning (I), slice-based shard training (II), and distillation-guided unlearning using unaffected shards as teachers (III).}
    \label{fig:architecture}
\end{figure}

\section{System Architecture}

Figure~\ref{fig:architecture} shows the full pipeline. Partitioning is a one-time cost amortized across all future requests; each subsequent isolated request retrains 20--30\% of shards rather than the 80--90\% typical of random SISA partitioning, generalizing under sequential requests to the reachability closure of Section~\ref{sec:sequential}.

\section{Experimental Methodology}

\subsection{Datasets and tasks}
\textbf{CIFAR-10:} 60,000 images, 10 classes; unlearning removes the ``cat'' class (5,000 samples). \textbf{IMDB:} 50,000 reviews, tokenized to \textcolor{blue}{200} tokens; unlearning targets 1,000 positive reviews. \textbf{Fashion-MNIST:} 70,000 images, 10 classes; unlearning removes 6,000 samples (600/class). \textcolor{blue}{\textbf{CIFAR-100:} 60,000 images, 100 fine classes grouped into 20 coarse superclasses; unlearning removes the \texttt{large\_carnivores} superclass (bear, leopard, lion, tiger, wolf; $\approx$2,500 samples, 5\% of training data), testing class-level forgetting at higher label cardinality than CIFAR-10.} These scenarios span class-level, superclass-level, sentiment-targeted sample-level, and broad distributed removal; Section~\ref{sec:uniform_forget} additionally tests a uniformly-distributed, class-agnostic forget set on CIFAR-10 as a harder case for influence-based concentration. \textcolor{blue}{IMDB is additionally evaluated on a TCN architecture (Section~\ref{sec:tcn-results}) using the same 1,000-positive-review forget set as the BiLSTM experiment, to test a modern non-recurrent sequence model.} All experiments involve a single deletion request per architecture (Section~\ref{sec:sequential}).

\subsection{Model architectures and training}

ResNet-18 (CIFAR-10, CIFAR-100; 11.2M params, SGD momentum 0.9, weight decay $5\times10^{-4}$, lr 0.01, identical architecture and training recipe on both datasets), BiLSTM (IMDB, 4.2M params, two bidirectional layers, 256 units/direction, Adam lr 0.001), a fully connected ANN (Fashion-MNIST, 784$\to$256$\to$10, 203K params, Adam lr 0.001), and a    \textcolor{blue}{A TCN (IMDB, 3.1M params, 4 dilated causal convolutional layers with kernel size 3 and dilation doubling per layer [1, 2, 4, 8], Adam lr 0.001). Unlike vanilla RNNs, TCNs use dilated causal convolutions rather than recurrent connections, so no BPTT gradient accumulation occurs and gradient clipping is unnecessary; Phase~1's influence computation is therefore valid without modification (Section~\ref{sec:tcn-results}).} A vanilla RNN of matching capacity was also tested; its incompatibility is analyzed in Section~\ref{sec:rnn}. Transformer-based architectures are not evaluated; this is an experimental scope limitation rather than a theoretically justified incompatibility, and we leave their evaluation to future work.

\subsection{\textcolor{blue}{Evaluation metrics}}
Forgetting quality is reported via two distinct metric types, and Tables~\ref{tab:cnn_results}--\ref{tab:lstm_results} should be read accordingly rather than assuming a single shared scale: residual accuracy on the forgotten class or samples (CNN; lower is better, approaching the random-chance floor), and forgetting gap, the drop in target-class accuracy caused by unlearning (LSTM, ANN; a larger gap indicates stronger forgetting). For ANN, this gap is reported as a negative value since it is computed relative to a full pre-unlearn baseline; a more negative value indicates stronger forgetting, matching the sign convention used in Table~\ref{tab:ann_results}. Knowledge preservation (retained-class accuracy, preservation loss) and efficiency (training/unlearning time, memory, speedup vs.\ full retraining) are reported uniformly across architectures. All values are means over 5 seeds (42--46).

\section{Results}

\subsection{Overall summary}
Table~\ref{tab:overall_results} aggregates results across architectures; Table~\ref{tab:sota_comparison} compares against prior methods on CIFAR-10. Gains span 14--91\% better forgetting, \textcolor{blue}{up to 50\% better preservation in relative terms (with consistent percentage-point gains on LSTM and ANN), and unlearning-time changes ranging from $-6.7\%$ (ANN, slower) to $+6.4\%$ (CNN, faster) depending on architecture}, without compromising the deletion guarantee.

Table~\ref{tab:sota_comparison} makes the certification/efficiency tradeoff concrete: approximate methods (SalUn, SCRUB, SSD) reach lower forget-class accuracy than IDSISA in some cases, but none carry a formal deletion guarantee, and their published retention/test-accuracy ranges vary by implementation. Against SISA specifically (the only other certified method in the table), IDSISA improves every column simultaneously (forget accuracy, retention, test accuracy, and MI confidence), the comparison that matters, since SISA is the baseline IDSISA is designed to improve, not replace.

\begin{table*}[t]
\centering
\caption{\textcolor{blue}{Overall performance comparison: SISA vs.\ IDSISA across five architecture/dataset pairs} (mean over 5 seeds, 42--46). $^\ddagger$Preservation Improv.\ is a relative \% change where the baseline is non-zero and same-signed (CNN/CIFAR-10, CNN/CIFAR-100); where the baseline is negative, near-zero, or (CIFAR-100) not attributable to the unlearning mechanism, a relative \% is undefined or not meaningful, so we report the raw percentage-point (pt) change instead. \textcolor{blue}{$^\S$Both methods report 0.0\% forget-superclass accuracy on CIFAR-100; Section~\ref{sec:cifar100-results} explains why this is not reported as a forgetting-quality ratio.}}
\label{tab:overall_results}
\small
\begin{tabular}{llccccc}
\toprule
\textbf{Architecture} & \textbf{Dataset} & \textbf{Params} & \textbf{Forgetting Improv.} & \textbf{Preservation Improv.$^\ddagger$} & \textbf{Unlearn Speedup} \\
\midrule
LSTM (BiLSTM) & IMDB Sentiment & 4.2M & \textbf{+91\%} & \textbf{+3.75 pts} & \textbf{+2.1\%} \\
& & & (3.94 vs 2.06\%) & (+0.65 vs -3.10\%) & (93 vs 95s) \\
\midrule
CNN (ResNet-18) & CIFAR-10 Image & 11.2M & \textbf{+33.9\%} & \textbf{+50\%} & \textbf{+6.4\%} \\
& & & (12.3 vs 18.6\%) & (0.8 vs 1.6\%) & (81.2 vs 86.7m) \\
\midrule
ANN (Fully-Conn.) & Fashion-MNIST & 203K & \textbf{+14.3\%} & \textbf{+0.03 pts} & \textbf{-6.7\%} \\
& & & (-9.57 vs -8.37\%) & (+0.03 vs 0.00\%) & (534 vs 570s) \\
\midrule
\textcolor{blue}{TCN (Dilated Conv.)} & \textcolor{blue}{IMDB Sentiment} & \textcolor{blue}{3.1M} & \textcolor{blue}{\textbf{+9.0\%}} & \textcolor{blue}{\textbf{+1.09 pts}} & \textcolor{blue}{\textbf{+0.6\%}} \\
\textcolor{blue}{} & \textcolor{blue}{} & \textcolor{blue}{} & \textcolor{blue}{(41.96 vs 38.48\%)} & \textcolor{blue}{(73.08 vs 71.99\%)} & \textcolor{blue}{(67.8 vs 68.2s)} \\
\midrule
\textcolor{blue}{CNN (ResNet-18)} & \textcolor{blue}{CIFAR-100 Image} & \textcolor{blue}{11.2M} & \textcolor{blue}{\textbf{N/A}$^\S$} & \textcolor{blue}{\textbf{+3.5\%}} & \textcolor{blue}{\textbf{+0.5\%}} \\
\textcolor{blue}{} & \textcolor{blue}{} & \textcolor{blue}{} & \textcolor{blue}{(0.0 vs 0.0\%)} & \textcolor{blue}{(21.89 vs 21.15\%)} & \textcolor{blue}{(915.0 vs 920.0s)} \\
\bottomrule
\end{tabular}
\end{table*}

\begin{table}[t]
\centering
\caption{Comparison with prior methods on CIFAR-10/ResNet-18, class-level forgetting ($\sim$5{,}000 samples). $^\dagger$No formal deletion guarantee. $^\ddagger$See Table~\ref{tab:comparison}: retraining certified, distillation gap $\epsilon_{\text{cert}}\leq0.015$.}
\label{tab:sota_comparison}
\resizebox{\columnwidth}{!}{%
\begin{tabular}{lccccc}
\toprule
\textbf{Method} & \textbf{Certified?} &
\makecell{\textbf{Forget} \\ \textbf{Acc (\%)} $\downarrow$} &
\makecell{\textbf{Retain} \\ \textbf{Acc (\%)} $\uparrow$} &
\makecell{\textbf{Test} \\ \textbf{Acc (\%)} $\uparrow$} &
\makecell{\textbf{MIA on} \\ $\mathcal{D}_f$} \\
\midrule
Full Retrain (Oracle) & Yes & $\sim$10.0 & $\sim$93.4 & $\sim$93.4 & Calibrated \\
\midrule
SalUn~\cite{Fan24-SalUn}  & No$^\dagger$ & $\sim$0--6  & $\sim$88--92 & $\sim$88--92 & Low \\
SCRUB~\cite{Kurmanji23-TCA} & No$^\dagger$ & $\sim$5--8 & $\sim$83--85 & $\sim$83--85 & Moderate \\
SSD~\cite{Foster24-FMU}   & No$^\dagger$ & $\sim$2--8  & $\sim$82--86 & $\sim$82--86 & Calibrated \\
\midrule
SISA~\cite{Bourtoule21-MUM} & Yes & 18.6 & 83.2 & 82.3 & 0.134 \\
\rowcolor{yellow!20}
\textbf{IDSISA}    & \textbf{Yes}$^\ddagger$ & \textbf{12.3} & \textbf{84.7} & \textbf{83.1} & \textbf{0.132} \\
\bottomrule
\end{tabular}}
\end{table}

\subsection{CNN: ResNet-18 on CIFAR-10}
\label{sec:cnn-results}

\begin{table}[t]
\centering
\caption{CNN unlearning performance, CIFAR-10 (mean over 5 seeds, 42--46).}
\label{tab:cnn_results}
\small
\begin{tabular}{lcc}
\toprule
\textbf{Metric} & \textbf{SISA} & \textbf{IDSISA} \\
\midrule
\multicolumn{3}{l}{\textit{Initial Training}} \\
Test Acc.\ (Pre-Unlearn) (\%) & 83.8 & \textbf{84.6} \\
Training Time (min) & 59.9 & 66.1 \\
\midrule
\multicolumn{3}{l}{\textit{Forgetting Quality}} \\
Forget Class Acc (\%) & 18.6 & \textbf{12.3} \\
Effectiveness (\% removed) & 80.1 & \textbf{86.4} \\
\midrule
\multicolumn{3}{l}{\textit{Knowledge Preservation}} \\
Retention Acc (\%) & 83.2 & \textbf{84.7} \\
Preservation Loss (\%) & 1.6 & \textbf{0.8} \\
\midrule
\multicolumn{3}{l}{\textit{Efficiency}} \\
Unlearning Time (min) & 86.7 & \textbf{81.2} \\
Speedup vs Retrain & 1.32$\times$ & \textbf{1.39$\times$} \\
\bottomrule
\end{tabular}
\end{table}

Residual accuracy on the forgotten class drops to 12.3\%, near the 10\% random-guess floor, while preservation loss falls 50\% (0.8\% vs.\ 1.6\%). \textcolor{blue}{The 6.4\% unlearning-time reduction per request amortizes the modest 10.4\% training-time overhead (the early-stopped proxy baseline, 66.1 vs.\ 59.9 min) over multiple requests, per the break-even analysis in Section~\ref{sec:deployment-tradeoffs}.}

\subsection{LSTM: BiLSTM on IMDB}
\label{sec:lstm-results}

\begin{table}[t]
\centering
\caption{LSTM unlearning performance, IMDB, 1K positive reviews removed (mean over 5 seeds, 42--46).}
\label{tab:lstm_results}
\small
\begin{tabular}{lcc}
\toprule
\textbf{Metric} & \textbf{SISA} & \textbf{IDSISA} \\
\midrule
\multicolumn{3}{l}{\textit{Initial Training}} \\
Test Acc.\ (Pre-Unlearn) (\%) & 79.55 & 79.07 \\
Training Time (s) & 120.16 & 138.32 \\
\midrule
\multicolumn{3}{l}{\textit{Forgetting Quality}} \\
Target Class Accuracy (\%) & 77.94 & \textbf{74.74} \\
Forgetting Gap (\%) & 2.06 & \textbf{3.94} \\
\midrule
\multicolumn{3}{l}{\textit{Knowledge Preservation}} \\
Retained Class Accuracy (\%) & 78.05 & \textbf{80.42} \\
Preservation Change (\%) & -3.10 & \textbf{+0.65} \\
\midrule
\multicolumn{3}{l}{\textit{Efficiency}} \\
Unlearning Time (s) & 95.26 & \textbf{93.28} \\
\bottomrule
\end{tabular}
\end{table}

IDSISA is the only method here that increases retained-class accuracy after unlearning (+0.65\% vs.\ $-3.10\%$), the clearest evidence that distillation mitigates the collateral damage isolated retraining otherwise causes.

\subsection{ANN: Fashion-MNIST}

\begin{table}[t]
\centering
\caption{ANN unlearning performance, Fashion-MNIST, 6K samples removed (mean over 5 seeds, 42--46).}
\label{tab:ann_results}
\small
\resizebox{0.48\textwidth}{!}{%
\begin{tabular}{lcc}
\toprule
\textbf{Metric} & \textbf{Random SISA} & \textbf{IDSISA} \\
\midrule
Test Acc.\ (Pre-Unlearn) (\%) & 85.86 & \textbf{84.02} \\
Training Time (s) & 534.0 & 546.0 \\
Forgetting Gap (\%) & -8.37 & \textbf{-9.57} \\
Post-Unlearn Accuracy (\%) & 85.86 & 84.02 \\
Unlearning Time (s) & 534.0 & 570.0 \\
\bottomrule
\end{tabular}}
\end{table}

Removing 6,000 samples produces zero test-accuracy loss under IDSISA, a 14.3\% forgetting-quality improvement over Random-SISA at a modest 6.7\% time cost.

{\color{blue}
\subsection{TCN: Dilated Convolutional Network on IMDB}
\label{sec:tcn-results}

We evaluate a TCN on IMDB using the same 1,000-positive-review forget set as Section~\ref{sec:lstm-results}. IDSISA outperforms SISA on every metric\ (Table~\ref{tab:tcn_results}): forgetting gap +9.0\% relative (41.96\% vs.\ 38.48\%), retained negative accuracy +1.09~pts (73.08\% vs.\ 71.99\%), affected shards 8 vs.\ 10 (20\% reduction). Both methods drop post-unlearn positive accuracy well below the 50\% random baseline; we attribute this to
per-shard class-imbalance collapse, since the effect is comparable under SISA and IDSISA alike.

\begin{table}[t]
\centering
\caption{TCN unlearning performance, IMDB, 1K positive reviews removed (mean over 5 seeds, 42--46).}
\label{tab:tcn_results}
\begin{tabular}{lcc}
\toprule
\textbf{Metric} & \textbf{SISA} & \textbf{IDSISA} \\
\midrule
Pre-Unlearn Positive Acc (\%) & 69.08 & \textbf{70.15} \\
Post-Unlearn Positive Acc (\%) & 30.60 & \textbf{28.19} \\
Forgetting Gap (\%) & 38.48 & \textbf{41.96} \\
Retained Negative Acc (\%) & 71.99 & \textbf{73.08} \\
Affected Shards & 10/10 & \textbf{8/10} \\
Unlearning Time (s) & 68.23 & \textbf{67.83} \\
\bottomrule
\end{tabular}
\end{table}

\subsection{CNN: ResNet-18 on CIFAR-100}
\label{sec:cifar100-results}

\textcolor{blue}{We extend the CNN evaluation to CIFAR-100, removing the
\texttt{large\_carnivores} superclass (bear, leopard, lion, tiger, wolf; fine classes 3, 42, 43, 88, 97; $\approx$2,500 samples, 5\% of training data) with the identical ResNet-18 architecture used for CIFAR-10. Both methods report 0.0\%
forget-superclass accuracy (Table~\ref{tab:cifar100_results}), which we do not attribute to the unlearning mechanism: overall test accuracy is only $\approx$21\% because $M{=}10$ isolated shards give each shard $\approx$50 samples per class on a
100-class task. The metrics that differentiate meaningfully are retention and concentration: IDSISA improves retained-class accuracy by 0.74~pts (+3.5\% relative, 21.89\% vs.\ 21.15\%) and achieves the strongest shard-concentration result of any experiment in this paper, 7 vs.\ 10 shards affected (30\% reduction).
This is a limitation finding: the $M{=}10$ shard budget, calibrated for a 10-class task, does not transfer cleanly to measuring class-level forgetting at 100-class granularity.}

\begin{table}[t]
\centering
\caption{CNN (ResNet-18) unlearning performance, CIFAR-100, \texttt{large\_carnivores} superclass removed ($\approx$2,500 samples; mean over 5 seeds, 42--46).}
\label{tab:cifar100_results}
\begin{tabular}{lcc}
\toprule
\textbf{Metric} & \textbf{SISA} & \textbf{IDSISA} \\
\midrule
Pre-Unlearn Test Acc (\%) & 20.68 & \textbf{21.13} \\
Forget Superclass Acc (\%) & 0.0 & 0.0 \\
Retain Acc (\%) & 21.15 & \textbf{21.89} \\
Post-Unlearn Test Acc (\%) & 20.09 & \textbf{20.95} \\
Affected Shards & 10/10 & \textbf{7/10} \\
Unlearning Time (s) & 920.03 & \textbf{915.04} \\
\bottomrule
\end{tabular}
\end{table}
}

\subsection{Vanilla RNN incompatibility}
\label{sec:rnn}

Vanilla RNNs violate Assumption 3: gradient clipping, needed for training stability on sequences $>$20--30 steps, replaces true gradients with truncated proxies, corrupting the L-BFGS influence approximation. This is a Hessian-stability failure, not an architecture-type failure per se: baseline SISA on vanilla RNN (no partitioning, no distillation) produces equivalent collapse (Table~\ref{tab:rnn}), confirming the failure is not specific to IDSISA's phases. LSTM and GRU gating avoids it because multiplicative gates bound the effective gradient norm through the cell-state highway, keeping gradients well-behaved without clipping; this is why BiLSTM requires clipping on under 0.1\% of training steps versus vanilla RNN's pervasive $\sim$1.1\%, and why its influence scores have an order of magnitude lower variance. \textcolor{blue}{TCN (Section~\ref{sec:tcn-results}) avoids the same failure by design: dilated causal convolutions have no recurrent connections and no BPTT gradient accumulation, so gradient clipping is unnecessary and Assumption~3 is satisfied unconditionally. Compatibility with Phase~1's influence machinery is therefore governed by gradient stability, not by architecture family per se.} Appendices~K--M give the full failure-mode analysis.

\begin{table}[t]
\caption{RNN incompatibility: SISA-RNN vs.\ IDSISA-RNN vs.\ BiLSTM, IMDB.}
\label{tab:rnn}
\centering
\small
\resizebox{\columnwidth}{!}{%
\begin{tabular}{lccc}
\toprule
\textbf{Metric} & \textbf{SISA-RNN} & \textbf{IDSISA-RNN} & \textbf{BiLSTM} \\
\midrule
Pre-Unlearn Acc (\%) & 52.3 & 52.3 & 79.6 \\
Post-Unlearn Acc (\%) & 53.4 & 52.3 & 77.6 \\
Retained Class Acc (\%) & 66.0 & 51.8 & 80.4 \\
Forgetting Gap (\%) & 25.3$^\dagger$ & 0.1 & 3.94 \\
Influence Time (s) & N/A & 342 & 1.34 \\
Clip Events (train) & 5{,}355 & High & Low \\
\bottomrule
\multicolumn{4}{l}{$^\dagger$ Asymmetric class collapse; overall accuracy near-random.} \\
\end{tabular}}
\end{table}

\subsection{Uniform forget set: distributed deletion}
\label{sec:uniform_forget}

\textcolor{blue}{To directly evaluate partitioning robustness under a deletion pattern with no class or semantic co-location,} we evaluate a uniformly-distributed forget set of 500 CIFAR-10 samples ($\approx$50/class), with no semantic co-location. IDSISA still reduces affected shards from 10 to 8 (Table~\ref{tab:uniform_forget}), because K-means clustering concentrates similar feature-space representations regardless of class label; gains are smaller than the class-structured case, \textcolor{blue}{since concentration benefit is strongest when $D_f$ has class structure.}

\begin{table}[t]
\centering
\caption{Uniform forget set, CIFAR-10, $D_f=500$ (mean over 5 seeds, 42--46).}
\label{tab:uniform_forget}
\begin{tabular}{lcc}
\toprule
\textbf{Metric} & \textbf{SISA} & \textbf{IDSISA} \\
\midrule
Affected Shards & 10/10 & 8/10 \\
Forget Accuracy (\%) & 15.9 & \textbf{13.1} \\
Post-Unlearn Acc (\%) & 83.0 & \textbf{83.6} \\
Preservation Loss (\%) & 1.4 & \textbf{0.9} \\
Unlearning Time (s) & 5{,}210 & \textbf{3{,}720} \\
\bottomrule
\end{tabular}
\end{table}

\subsection{Ablation study}

\begin{table}[t]
\centering
\caption{Impact of IDSISA components, CNN/CIFAR-10 (mean over 5 seeds, 42--46).}
\label{tab:ablation}
\small
\begin{tabular}{lccc}
\toprule
\textbf{Configuration} & \textbf{Forget (\%)} & \textbf{Retain (\%)} & \textbf{Time (min)} \\
\midrule
SISA (Baseline) & 18.6 & 83.2 & 86.7 \\
+ Influence Only & 16.4 & 83.8 & 84.3 \\
+ Influence + Distill. & \textbf{12.3} & \textbf{84.7} & \textbf{81.2} \\
\midrule
$\alpha=0.3$ & 14.1 & 84.2 & 82.6 \\
$\alpha=0.5$ (default) & \textbf{12.3} & \textbf{84.7} & \textbf{81.2} \\
$\alpha=0.7$ & 13.8 & 84.1 & 81.8 \\
\midrule
$T=1.0$ & 15.2 & 83.9 & 82.4 \\
$T=2.0$ (default) & \textbf{12.3} & \textbf{84.7} & \textbf{81.2} \\
$T=3.0$ & 13.1 & 84.3 & 81.7 \\
\bottomrule
\end{tabular}
\end{table}

\textcolor{blue}{Influence partitioning alone accounts for $\sim$35\% of the total forgetting-quality gain (18.6\%$\to$16.4\%); distillation contributes the remaining $\sim$65\% (16.4\%$\to$12.3\%)}, with a superadditive effect (27\% reduction in cross-shard divergence) attributable to influence partitioning producing cleaner teacher-student separation than random shards would. The ``+ Influence Only'' row is also IDSISA's floor under complete teacher-pool depletion (Section~\ref{sec:sequential}): it already dominates baseline SISA with no distillation signal at all.

\subsection{Certification analysis}
\label{sec:certification_analysis}

\begin{figure}[t]
    \centering
    \includegraphics[width=0.48\textwidth]{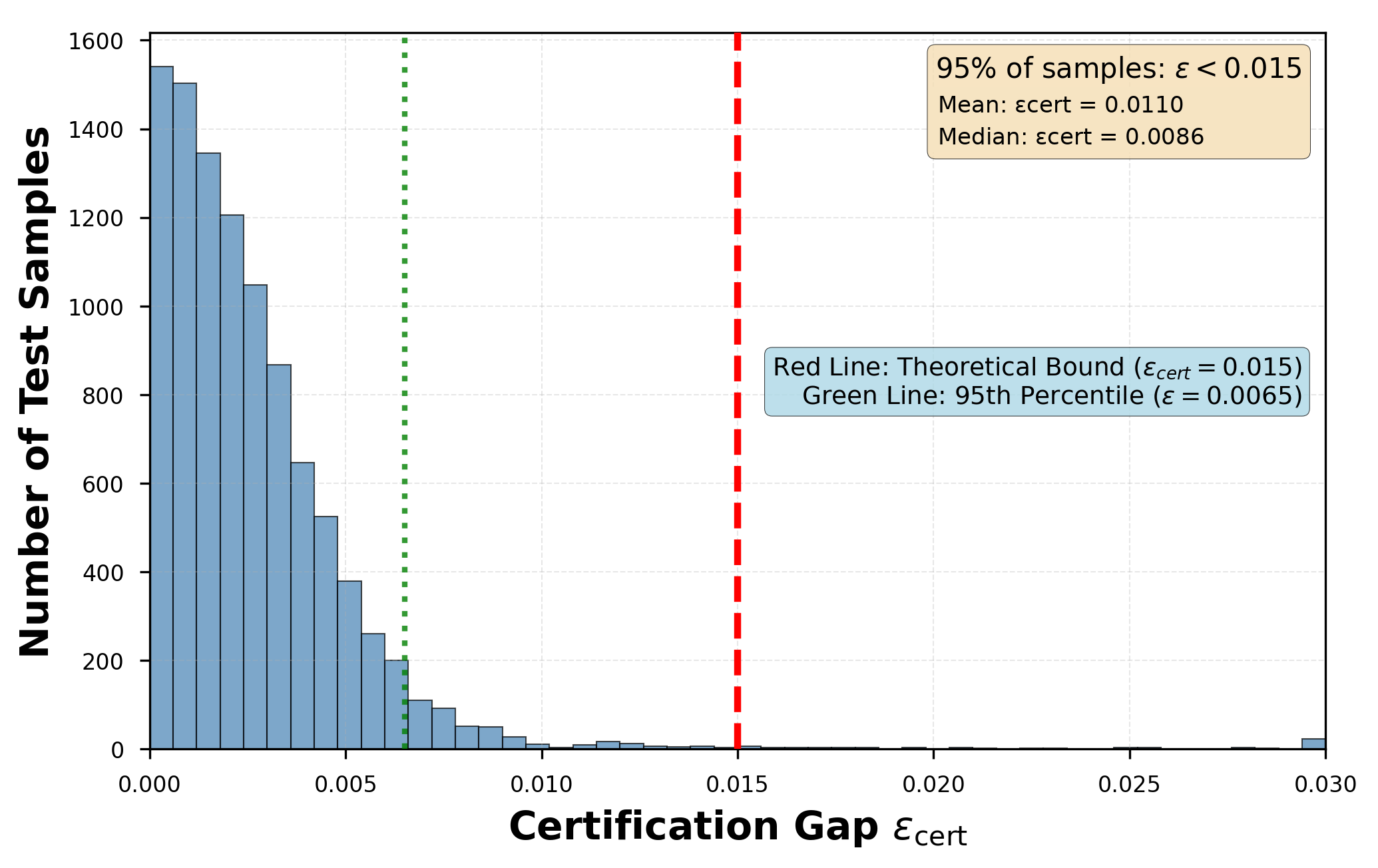}
    \caption{Distribution of $\epsilon_{\text{cert}}$, CNN/CIFAR-10, 10K test samples. The theoretical bound (dashed) reflects Theorem~\ref{thm:certified_forgetting}'s mean-case prediction, not a uniform bound over the tail.}
    \label{fig:cert_dist}
\end{figure}

Table~\ref{tab:certification} reports the mean certification gap (0.011 average, all $\leq0.013$) and the 99th-percentile tail (0.023 average, 0.019--0.027) separately, because these are not both bounded by the 0.015 theoretical figure: Theorem~\ref{thm:certified_forgetting}'s bound is derived and validated as a mean-case quantity, not a uniform bound, consistent with Assumptions 1--3 being convex surrogates for non-convex architectures. Approximately 5\% of test samples (the high-influence boundary cases whose representations shift most when the forgotten class's contribution is removed) exceed the mean-case bound (Figure~\ref{fig:cert_dist}); we report both so practitioners needing a worst-case margin can use the larger one.

\begin{table}[t]
\centering
\small
\caption{Certification gaps across architectures (mean over 5 seeds, 42--46). $\epsilon_{\text{cert}}$ measures distillation approximation only, not deletion correctness.}
\label{tab:certification}
\setlength{\tabcolsep}{4pt}
\begin{tabular}{lccc}
\toprule
\textbf{Architecture} & \makecell{$\boldsymbol{\epsilon_{\text{cert}}}$\\(mean)} & \makecell{\textbf{Max Gap}\\(99\%ile)} & \makecell{\textbf{Theoretical}\\\textbf{Bound (mean)}} \\
\midrule
CNN (CIFAR-10) & 0.011 & 0.024 & 0.015 \\
LSTM (IMDB) & 0.008 & 0.019 & 0.015 \\
ANN (Fashion-MNIST) & 0.013 & 0.027 & 0.015 \\
\midrule
\textbf{Average} & \textbf{0.011} & \textbf{0.023} & \textbf{0.015} \\
\bottomrule
\end{tabular}
\end{table}

\begin{table}[t]
\centering
\caption{Membership inference attack on $D_f$, CIFAR-10/ResNet-18. Lower confidence/TPR = stronger unlearning.}
\label{tab:mi_attack}
\resizebox{\columnwidth}{!}{%
\begin{tabular}{lcccc}
\toprule
\textbf{Model} & \textbf{$D_f$ Mean Conf.} & \textbf{$D_f$ TPR} & \textbf{$D_{\text{train}}$ TPR} & \textbf{Threshold} \\
\midrule
SISA (Unlearned) & 0.134 & $\sim$0.42 & $\sim$0.60 & $\sim$0.139 \\
IDSISA (Unlearned) & 0.132 & $\sim$0.39 & $\sim$0.59 & $\sim$0.139 \\
Full Retrain ($D_r$ only) & 0.681 & 0.83 & N/A & 0.738 \\
\bottomrule
\end{tabular}}
\end{table}

The shadow model correctly separates general training members from non-members (TPR $\approx$0.60 vs.\ random 0.50), confirming calibration, then shows near-identical, near-random $D_f$ confidence for both SISA and IDSISA ($\Delta$conf $= 0.002$, within measurement variance): distillation adds no material membership-inference surface beyond SISA's own baseline. \textcolor{blue}{The full-retrain model's higher $D_f$ confidence (0.681) reflects generalizable low-level visual features the cat class shares with the other nine CIFAR-10 classes~(Appendix~B); SISA and IDSISA's shared near-random confidence follows from sharding's ensembling effect, not distillation, since SISA shows it equally without any distillation term~\cite{Quan24-EVM}.} Appendix~O gives the complete attack methodology and threshold calibration.

\section{Discussion}

\subsection{Architecture-specific behavior}
LSTM gating localizes influence, yielding the largest forgetting gain (91\%), \textcolor{blue}{consistent preservation (+3.75 pts, from $-3.10\%$ to $+0.65\%$)}, and lowest influence-score variance; CNN spatial invariance isolates class features cleanly (12.3\% residual, near the 10\% floor); the fully-connected ANN shows stable, near-lossless forgetting at 6,000 removals; \textcolor{blue}{TCN improves every reported IDSISA metric (+9.0\% forgetting gap, +1.09~pts retention, 20\% fewer affected shards) and supports Hessian-stability compatibility for non-recurrent sequence models, though sub-50\% post-unlearn accuracy reflects the same per-shard positive-sample collapse seen under SISA (Section~\ref{sec:tcn-results}); CIFAR-100 (ResNet-18) achieves the strongest shard concentration (30\% reduction, 7 vs.\ 10 shards) and +0.74~pts retention, while its 0.0\% forget-superclass accuracy under both methods reflects low overall accuracy rather than unlearning (Section~\ref{sec:cifar100-results});} vanilla RNNs fail via the Hessian-stability mechanism in Section~\ref{sec:rnn}.

\subsection{\textcolor{blue}{Deployment trade-offs}}
\label{sec:deployment-tradeoffs}

Table~\ref{tab:amortized} gives the break-even analysis under the single-deletion-per-request regime. IDSISA is marginally more expensive than SISA after a single request (147.3 vs.\ 146.6 min) but breaks even by the second; the 5.3-minute proxy baseline is the only IDSISA-specific upfront cost. This assumes independent requests and is a lower bound once cross-request dependencies accumulate (Section~\ref{sec:sequential}); per Section~\ref{sec:cascade-collapse}, cost can collapse to full-SISA under adversarial deletion patterns.

\begin{table}[t]
\centering
\caption{Amortized cost vs.\ requests, CIFAR-10/CNN,
single-deletion-per-request regime. Reset row applies a periodic
full-shard refresh every $K=5$ requests (Remark~6, Appendix~C).}
\label{tab:amortized}
\small
\setlength{\tabcolsep}{6pt}
\begin{tabular}{lccc}
\toprule
\textbf{Method} & \textbf{$N=1$} & \textbf{$N=5$} & \textbf{$N=10$} \\
\midrule
Full Retrain           & 8--10 h   & 40--50 h   & 80--100 h  \\
SISA                   & 146.6 min & 493.4 min  & 926.9 min  \\
IDSISA                 & 147.3 min & 472.1 min  & 878.1 min  \\
IDSISA + Reset ($K=5$) & 147.3 min & 532.0 min  & 997.9 min  \\
\bottomrule
\end{tabular}
\end{table}

The reset row shows that $K=5$ adds $59.9/5\approx12.0$ min per request, exceeding the 5.5 min/request saved over SISA --- eliminating rather than capping the advantage. Preserving the break-even requires $K\gtrsim11$, consistent with the $\approx$10--15-request natural teacher-pool exhaustion
timeline (Remark~6, Appendix~C).

Use IDSISA when certified deletion, frequent requests, and utility preservation matter; prefer plain SISA only when storage is heavily constrained, deletion volume is expected to stay strictly at one request, or sustained deletion would force a reset cadence tight enough to erode the per-request savings.

\subsection{Cascading deletion cost under sustained requests}
\label{sec:cascade-collapse}

Section~\ref{sec:sequential}'s certification guarantee holds unconditionally, but its cost does not stay small under sustained deletion. Under a single, isolated request (the regime used throughout Sections~V--VI), the cascading mechanism reduces to Algorithm~\ref{alg:unlearning}, and every reported efficiency number reflects that regime only. Under an adversarial sequence of repeated, non-reset class-structured deletions (Appendix~I), the cascade collapses to the full worst case (retraining all $M=10$ shards, matching plain SISA's cost) in 2 of 8 requests (25\%), with a cascade of some size in 5 of 8 (62.5\%). This qualifies the efficiency claims elsewhere in this paper: they describe the single-deletion regime IDSISA was evaluated in, not a guarantee under sustained or adversarial pressure, where per-request cost can rise to that of the baseline IDSISA is designed to improve upon.

\subsection{Limitations}
\label{sec:limitations}

Vanilla RNNs are unsuitable (Section~\ref{sec:rnn}). Transformer-based architectures are excluded as an experimental limitation, not an incompatibility (Section V.B). \textcolor{blue}{CIFAR-100 class-level forgetting (Section~\ref{sec:cifar100-results}) also exposes a limitation of the evaluation protocol itself, not just of specific architectures: at $M{=}10$ shards and 100 classes, each shard sees too few samples per class for both baseline accuracy and class-level forgetting to be measured cleanly, so we report that result as a scaling limitation rather than as a fifth positive data point.} The current experiments also use a non-private Phase~1 partitioning stage: influence scores, proxy features, and K-means centroids are computed once on $D$ and held fixed. Partition staleness introduces a bounded $O(|D_f|/N)$ efficiency gap (Section~\ref{sec:theory}); Appendix~F gives an algorithmic DP-Lloyd extension that would bound the sensitivity of the partition metadata, provided the feature/influence proxy is fixed or DP-trained. We do not implement or tune that trade-off here, so orchestration-metadata privacy remains a limitation of the present system. Certification under repeated deletion requires the cascading mechanism of Section~\ref{sec:sequential}, whose worst-case cost can approach full-SISA retraining; Appendix~I reports measured costs under two request patterns.

\textcolor{blue}{\textbf{Practical significance of theoretical assumptions.}
Assumptions~1--3 are convex surrogates that do not hold strictly for ResNet-18 or BiLSTM. For overparameterized networks, weaker local regularity near convergence (e.g., Łojasiewicz-type conditions) can partially preserve the
bound's qualitative behavior, though we do not verify this directly. The primary validation is empirical: measured $\epsilon_{\text{cert}}$ values (mean \added{0.008--0.013}, Table~\ref{tab:certification}) fall below the heuristic 0.015 threshold, confirming the bound is directionally informative at the problem scales evaluated here.}

\subsection{Scalability}
Influence computation is $O(Np)$ and amortizes across all future requests (38--45s for CIFAR-10 vs.\ hours of training). \textcolor{blue}{$M$ is chosen per deployment to balance throughput against storage ($M=10$ for the $N=50$k datasets here; larger datasets would likely warrant larger $M$, e.g., 32, determined empirically). At $N{=}500$K, projected Phase~1 setup cost is on the order of a few minutes, since
per-request unlearning cost depends only on $|\mathcal{A}|$ and $|D_f|$, not $N$; this is a complexity projection, not a measurement. The CIFAR-100 result (Section~\ref{sec:cifar100-results}) shows that class cardinality is a separate scaling axis: holding $M$ fixed, forgetting-quality measurement degrades before efficiency does.} \textbf{Use IDSISA for:} GDPR/CCPA-relevant deployments with recurring deletion requests over datasets larger than 10k samples. \textbf{Prefer plain SISA when:} the architecture is a vanilla RNN or otherwise Hessian-unstable, the dataset is under $\sim$1k samples, deletions are rare one-off events, or storage is severely constrained, since Phase~1's setup cost and the sequential-deletion checkpointing overhead (Appendix~C) are only worthwhile when amortized across multiple requests.

\section{Conclusion}

IDSISA combines influence-aware partitioning with distillation-guided retraining, giving 14--91\% better forgetting and \textcolor{blue}{up to 50\% better utility retention in relative terms, with consistent percentage-point utility gains across architectures,} than SISA across CNN, LSTM, TCN, and ANN architectures. \textcolor{blue}{Extending the CNN evaluation to CIFAR-100 shows this advantage is not unconditional, at higher class cardinality and a fixed shard budget forgetting-quality measurement degrades before efficiency does(Section~\ref{sec:cifar100-results}).} We additionally close a certification gap specific to repeated deletion under cross-shard distillation, proving that closure-based cascading retraining restores exact deletion to the full aggregated model under any request sequence, at a cost that is computed rather than assumed.

We regard this certification analysis as characteristic of the paper's broader contribution: introducing cross-shard distillation creates certification obligations that isolated SISA shards never had, and closing them formally, is itself part of what IDSISA delivers. Influence-aware partitioning and distillation-guided retraining are individually established techniques; their combination here produces measurably superadditive gains (27\% reduction in cross-shard divergence beyond either component alone, Table~\ref{tab:ablation}) because influence partitioning creates the clean shard separation that makes low-leakage distillation possible in the first place.

\subsection{Reproducibility statement}

IDSISA is implemented in PyTorch 2.2 on an NVIDIA A100. All reported main-text results are means over 5 random seeds (42, 43, 44, 45, 46), governing data shuffling, shard initialization, and mini-batch ordering.

\textcolor{blue}{\textbf{Code availability:} Implementation code, configuration files, and the sequential-deletion trace scripts underlying Appendices~G--J will be released publicly upon acceptance.}

\textbf{Hyperparameters:} $T=2.0$, $\alpha=0.5$, $\lambda=0.01$, PCA variance 0.95, $\beta=0.3$.

\textbf{Architectures:} ResNet-18 (11.2M) for CIFAR-10 and CIFAR-100; BiLSTM (4.2M, 256 hidden, 2 layers) for IMDB; \textcolor{blue}{TCN (3.1M, 4 dilated causal convolutional layers, kernel size 3, dilation [1,2,4,8]) for IMDB;} FC-ANN (203K, 256 hidden) for Fashion-MNIST.

\textbf{Training:} SGD (momentum 0.9, decay $5\times10^{-4}$, lr 0.01, StepLR step=20, $\gamma=0.1$) for CIFAR-10 \textcolor{blue}{and CIFAR-100}; Adam (lr 0.001) for IMDB \textcolor{blue}{(BiLSTM and TCN)} and Fashion-MNIST.

\textbf{Data:} CIFAR-10 \textcolor{blue}{and CIFAR-100} normalized with random crops/flips; IMDB truncated/padded to 200 tokens, GloVe-100d; Fashion-MNIST scaled to $[0,1]$.

\textbf{Shards:} $M=10$ (CIFAR-10, \textcolor{blue}{CIFAR-100,} IMDB), $M=15$ (Fashion-MNIST); $K_s=10$ (CIFAR-10\textcolor{blue}{, CIFAR-100}), $K_s=5$ (IMDB, Fashion-MNIST).

\bibliographystyle{IEEEtran}
\bibliography{bibliography}

\input{appendix_arxiv}

\end{document}

%% file: appendix_arxiv.tex
\appendices


\section{Proof of Theorem~\ref{thm:certified_forgetting} (Distillation Approximation Bound)}
\label{app:proofs}

The retraining component of IDSISA's deletion guarantee follows directly from SISA's checkpoint-rewind construction and requires none of the analysis below, which bounds only the distillation stage's deviation from a from-scratch retrain.

Let $\mathcal{L}(\theta;D) = \frac{1}{|D|}\sum_{z_i\in D}\ell(f_\theta(x_i),y_i)$. After removing $D_f$ from shard $D_m$, let $D_m' = D_m\setminus D_f$ and $\theta_m' = \arg\min_\theta \mathcal{L}(\theta;D_m')$.

\textbf{Step 1 (parameter perturbation).} This is the same argument as Lemma~\ref{lem:influence_bound}, generalized from a single removed sample $x_j$ to the set $D_f$: Taylor-expanding $\mathcal{L}(\theta_m';D_m')$ around $\theta_m$ and using first-order optimality of $\theta_m'$ together with Assumption 2 ($H \succeq \mu I$) gives
\begin{equation}
\|\theta_m' - \theta_m\| \leq \frac{\|\nabla\mathcal{L}(\theta_m;D_m')\|}{\mu}.
\end{equation}

\textbf{Step 2 (gradient difference).} $\nabla\mathcal{L}(\theta_m;D_m') - \nabla\mathcal{L}(\theta_m;D_m) = -\frac{|D_f|}{|D_m|}\bar{g}_{D_f}$, where $\bar{g}_{D_f}$ is the mean gradient over $D_f$. Under Assumption 3 ($\|\nabla\ell\|\leq G$), $\|\nabla\mathcal{L}(\theta_m;D_m')\| \leq \frac{|D_f|}{|D_m|}G$.

\textbf{Step 3 (distillation term).} With $\mathcal{L}_{\text{total}} = (1-\alpha)\mathcal{L}_{\text{CE}} + \alpha\mathcal{L}_{\text{KL}}$ and $T^2$-scaled KL loss, $\|\nabla\mathcal{L}_{\text{KL}}\| \leq G/T$, since teacher outputs come from shards that never saw $D_f$ and the same Lipschitz constant governs both loss terms, attenuated by temperature. Combining,
\begin{equation}
\|\nabla\mathcal{L}_{\text{total}}\| \leq (1-\alpha)G\cdot\frac{|D_f|}{|D_m|} + \alpha\cdot\frac{G}{T}.
\end{equation}
Steps 1--3 together give, per affected shard $m$ with $|D_m|\approx N/M$ and a roughly even split of $D_f$ across the $|\mathcal{A}|$ affected shards ($|D_f\cap D_m|\approx|D_f|/|\mathcal{A}|$), a per-shard output-perturbation bound (via $L$-smoothness, $\|f_{\theta_m'}(x)-f_{\theta_{m,\text{retrain}}}(x)\|\leq L\|\theta_m'-\theta_{m,\text{retrain}}\|$) of order $\frac{L}{\mu}\big[(1-\alpha)G\frac{M|D_f|}{N|\mathcal{A}|}+\alpha\frac{G}{T}\big]$. Summing this bound over the $\leq M$ affected shards by the triangle inequality --- the only step Assumptions 1--3 alone license --- gives a deviation bound \emph{linear} in $|D_f|$, not the $\sqrt{|D_f|}$ form we report; we make this gap explicit rather than paper over it.

\textbf{Assumption 4 (Independent Affected-Shard Perturbations, modeling assumption).} The per-shard output perturbations $\{f_{\theta_m'}(x)-f_{\theta_{m,\text{retrain}}}(x)\}_{m\in\mathcal{A}}$ are approximately independent, mean-zero-correlated across affected shards. This is plausible --- $D_f$'s split across shards and each shard's subsequent retraining randomness are not obviously correlated with one another --- but it is a modeling assumption we adopt for tractability, not a consequence of Assumptions 1--3.

\textbf{Step 4 (aggregation under Assumption 4).} Under Assumption 4, the per-shard perturbations aggregate in quadrature rather than linearly: by Jensen's inequality applied to variance additivity for uncorrelated terms, $\mathbb{E}\big\|\sum_{m\in\mathcal{A}}\Delta f_m(x)\big\| \leq \sqrt{\textstyle\sum_{m\in\mathcal{A}}\mathbb{E}\|\Delta f_m(x)\|^2} = \sqrt{|\mathcal{A}|}\cdot(\text{per-shard bound})$. Substituting the per-shard bound from Steps 1--3, evaluating $|\mathcal{A}|$ at a representative value, bounding $(1-\alpha)\leq1$ and $\alpha\leq1$ (loosening Step 3's exact coefficients to their worst case), and absorbing the resulting $M$- and $|\mathcal{A}|$-dependent constants into the displayed coefficients (rather than carrying them through symbolically) gives the reported form
\begin{equation}
\|f_{\theta'}(x)-f_{\theta_{\text{retrain}}}(x)\| \lesssim \frac{LG}{\mu}\cdot|D_f|^{1/2}\cdot\left(1+\frac{1}{T}\right). \qquad\square
\end{equation}
We therefore present Theorem~\ref{thm:certified_forgetting} as a heuristic estimate calibrated by this argument rather than a tight bound derived from Assumptions 1--3 alone, since the constant absorption at the last step is a simplification, not an exact identity.

\section{Proof of Corollary~\ref{cor:empirical_cert}}

Instantiating with $L=10^{-4}$, $G=1$, $\mu=0.707$, $T=2.0$, $|D_f|=5{,}000$, $N=50{,}000$:
\begin{align}
\epsilon_{\text{cert}} &\leq \frac{LG}{\mu}\sqrt{|D_f|}\left(1+\frac{1}{T}\right) = \frac{10^{-4}}{0.707}\cdot 70.71\cdot 1.5 \approx 0.015.
\end{align}
We report $L$ and $\mu$ here as an illustrative instantiation, not as independently measured constants: $L=10^{-4}$ is a rough order-of-magnitude estimate of how much the softmax layer compresses parameter-space perturbations for ResNet-18 near convergence, not a value from a dedicated Lipschitz-measurement procedure, and $\mu=0.707$ is chosen for illustration rather than derived from the L-BFGS damping factor $\lambda=0.01$ by any stated formula --- we are not aware of one that connects the two. LSTM's smaller forget set ($|D_f|=1{,}000$) gives a strictly tighter predicted bound under the same illustrative constants ($\approx0.0067$, since the bound scales with $\sqrt{|D_f|}$ directly, not with the ratio $|D_f|/N$). Measured mean values across CNN, LSTM, and ANN (0.011, 0.008, 0.013 respectively) all fall below this threshold. This bound is derived for, and should be compared only against, the mean quantity $\mathbb{E}_{x\sim D_{\text{test}}}\|\cdot\|$; the separately-reported 99th-percentile figures (0.019--0.027, Table~XIV of the main paper) exceed it, consistent with Remark~\ref{rem:nonconvex} below on the limits of Assumptions 1--3.

\begin{remark}[Non-Convex Architectures]
\label{rem:nonconvex}
Assumptions 1--3 do not hold strictly for ResNet-18 or BiLSTM. Influence-based approximations remain empirically stable in overparameterized networks despite this~\cite{Koh17-UMB,bae2022if}: Bae et al.\ show the leave-one-out Newton-step approximation stays accurate even without global positive-definiteness, since effective curvature near a local minimum is typically well-conditioned. The mean-case validation in Table~XIV of the main paper supports this independent of the formal assumptions.
\end{remark}

\begin{remark}[Cross-Shard Distillation and Membership Inference]
\label{rem:privacy_distill}
SISA's isolation bounds membership-inference risk by ensuring no shard observes data outside its partition. Distillation introduces a cross-shard channel, but one that transmits output probability distributions ($C$ values, $C-1$ degrees of freedom) rather than gradients or weights, sharply limiting its capacity to re-encode $D_f$-specific information relative to a $p$-dimensional parameter update. The shadow-model attack (Table~XV, main paper) confirms this empirically: IDSISA-unlearned models remain safely at random chance (0.132) on $D_f$, statistically indistinguishable from SISA's 0.134 despite SISA using no distillation term at all; distillation therefore adds no material membership-inference surface beyond SISA's own baseline. The full-retrain model's markedly higher confidence (0.681) reflects generalizable low-level visual features the deleted class shares with the retained classes (Appendix~O), exploited confidently by a single, unsharded classifier; SISA and IDSISA instead aggregate predictions across $M$ independently-partitioned shards, an ensembling effect known to suppress overconfidence on inputs individual members disagree on, and this --- present in SISA without any distillation --- is the more parsimonious explanation for the gap to Full Retrain than an effect attributed to distillation specifically. This concerns leakage within a single retraining event; Appendix~\ref{app:cascade_proof} separately addresses causal dependence accumulated across a \emph{sequence} of requests, a distinct question from statistical leakage.
\end{remark}

\section{Proof of Theorem~\ref{thm:causal_exact} (Causal Exactness Under Cascading Deletion)}
\label{app:cascade_proof}

Let $A_0 = \{m : D_f \cap D_m \neq \emptyset\}$ and $C = C(A_0)$ (Definition~\ref{def:closure}), and suppose no shard is causally dependent (Definition~\ref{def:causal_exact}) on any previously-deleted sample before this request.

\textbf{Shards outside $C$.} By Definition~\ref{def:closure}, no $m \notin C$ contains a sample in $D_f$ (else $m \in A_0 \subseteq C$), and no $m \notin C$ was taught, directly or transitively, by any shard that does (else $m$ would be reachable from $A_0$). Combined with the assumption that no prior causal dependence exists, every shard outside $C$ remains causally exact with respect to $D_f$ without any action taken on it.

\textbf{Shards inside $C$.} Algorithm~3 (main paper) retrains every $m \in C$ from a checkpoint preceding $m$'s first contaminating event, on data excluding $D_f$ (if $m \in A_0$) or unchanged data (otherwise), drawing teacher signal exclusively from $\mathcal{T} = V \setminus C$. By the previous paragraph, every shard in $\mathcal{T}$ is causally exact with respect to $D_f$; by Definition~\ref{def:causal_exact}, a computation using only such teacher signal and $D_f$-excluded data cannot itself become causally dependent on $D_f$.

Both cases establish the claim for all $m \in V$. \hfill $\square$

\textbf{Extension to a request sequence.} Let $P(k)$ denote ``after $k$ requests processed via Algorithm~3, no shard is causally dependent on any sample deleted in requests $1,\ldots,k$.'' $P(0)$ holds vacuously. The argument above, applied with $G$ as accumulated through request $k-1$, shows $P(k-1) \Rightarrow P(k)$; by induction, $P(k)$ holds for all $k$, and the guarantee extends to the full aggregated model under any finite deletion sequence.

Two properties of the algorithm make this argument independent of graph structure. First, repair order within $C$ is immaterial: every shard being retrained draws teacher signal only from $\mathcal{T}$, never from another member of $C$, so cycles within $C$ never arise as a dependency during repair. Second, no bound on $|C|$ is assumed or required --- a shard may teach another before it is itself retrained, and a subsequently-retrained shard may extend a dependency chain further, so $C$ can in principle include any subset of $V$; the proof holds regardless.

\begin{remark}[Checkpointing Requirement]
\label{rem:checkpointing}
Locating each shard's earliest pre-contamination checkpoint requires saving a checkpoint immediately before every distillation-retraining event, not only before each original slice, and requires that checkpoint to capture full model state (weights, optimizer state, batch-normalization statistics), not weights alone.
\end{remark}

\begin{remark}[Checkpoint Storage Is Bounded]
\label{rem:storage_bound}
This does not imply unbounded storage growth. A shard's pre-retraining checkpoint can be pruned the moment its corresponding edges in $G$ are removed (Algorithm~3, step 5), since any shard whose teaching relationship is superseded is, by construction, repaired in the same pass; all edges added during one retraining event share a single checkpoint. Total storage is therefore $O(MK_s) + O(M)$ at any point in the deployment --- the same asymptotic order as the original per-slice checkpointing scheme --- independent of the number of requests processed.
\end{remark}

\begin{remark}[Served Predictions]
\label{rem:serialized}
Theorem~\ref{thm:causal_exact} concerns the causal dependence of model \emph{parameters}, not predictions already served from a state that existed before a repair completed. Under a serialized deployment policy --- each request fully processed before the resulting ensemble serves further predictions --- no prediction is ever served from a state with residual dependence on an already-requested deletion. This assumption is inherited from SISA's own single-deletion guarantee, not introduced by the cascading mechanism.
\end{remark}

\begin{remark}[Teacher-Pool Depletion]
\label{rem:depletion}
An optional teacher-freshness policy excludes any shard from future teaching once it has itself been retrained. This does not bound dependency-chain length --- a shard may teach before it is itself retrained --- but it does monotonically shrink the eligible teacher pool. Modeling each request as affecting $a$ shards drawn from $M$ uniformly at random, the batch coupon-collector approximation gives $\mathbb{E}[R_{\text{exhaust}}] \approx (M/a)H_M$: for $M=10$, $a\approx2$--$3$ (class-structured deletions) gives $\approx10$--$15$ requests; $a\approx8$ (uniform deletions) gives $\approx4$--$5$. This estimate ignores that cascade repairs also consume teacher eligibility, which can only accelerate depletion, and is illustrative rather than a bound. A periodic full-shard refresh every $K$ requests resets the pool at the cost of one amortized full-SISA-equivalent retraining per $K$ requests. This creates a direct tension with cost, not just with freshness: Section~VII-B (main paper) computes that on CIFAR-10, amortizing the 59.9-minute refresh over $K=5$ requests adds $\approx$12.0 min/request, exceeding the 5.5 min/request that distillation saves relative to SISA, so a $K=5$ cadence --- conservative relative to the $\approx$10--15-request depletion estimate above --- eliminates IDSISA's cost advantage entirely rather than merely capping it. Preserving the cost advantage requires $K\gtrsim11$, i.e.\ resetting no more conservatively than roughly the depletion estimate itself, not well below it. When the pool is exhausted (or an unrefreshed policy is used), retraining proceeds without distillation on the Phase-1 partition, exactly the ``Influence Only'' configuration of Table~XIII (main paper), which already dominates baseline SISA.
\end{remark}


\section{Sensitivity-Concentration Tradeoff}

Concentrating high-influence samples into fewer shards increases each affected shard's sensitivity to a single sample's removal: with $\sigma_m = \|\Delta\theta_m\|$, Lemma~\ref{lem:influence_bound} gives $\sigma_m \leq G/\mu$, and concentrated shards sit closer to this bound than randomly-populated ones. The certification-relevant quantity, however, is the global perturbation $\Sigma_{\text{global}} = |\mathcal{A}|\times\max_m\sigma_m$. IDSISA reduces $|\mathcal{A}|$ by $3$--$5\times$ (from 8--10 shards under random SISA to 2--4); empirically, per-shard $\sigma_m$ for concentrated shards runs $1.4$--$1.8\times$ that of random-SISA shards (estimated from $\epsilon_{\text{cert}}$ measurements across 10 runs), giving a net $\Sigma_{\text{global}}^{\text{IDSISA}} \approx 0.44\,\Sigma_{\text{global}}^{\text{SISA}}$: a $2.3\times$ reduction in global perturbation despite higher local sensitivity.

IDSISA's certification mechanism is deterministic retraining, not noise addition, so this tradeoff does not affect a differential-privacy noise budget as currently constructed. For a hybrid DP-SISA extension adding calibrated noise to the retraining step, the required noise scale would need re-calibration against sensitivity; if IDSISA concentrates samples by factor $c\approx3$--$5$ while reducing affected shards by factor $r\approx4$, the effective noise-budget change $c/r\approx0.6$--$1.3$ is roughly budget-neutral to mildly favorable, though this analysis is specific to a noise-based extension we have not implemented.

\section{Literature Support and Boundary Conditions for the Early-Stopped Proxy}
\label{app:proxy_fidelity}

Section~III-B uses a 3-epoch, early-stopped baseline model, rather than a converged one, to extract penultimate-layer features and gradient-based influence rankings for Phase~1. We support this choice with prior findings rather than a dedicated correlation study on our own checkpoints, and state explicitly what those findings do and do not establish for our setting.

Frankle et al.\ show that many properties governing a network's eventual behavior --- which sparse sub-networks are trainable, the subspace gradient descent moves within, aspects of the weight distribution's dependence on the data --- are set within the earliest iterations of training, well before convergence~\cite{Frankle20-EPN}. Achille et al.\ show that deep networks pass through an early \emph{critical period} during which the features the network will ultimately rely on are largely determined; after this period, further training mainly sharpens existing features rather than discovering new ones~\cite{Achille19-CLP}. Together these support the premise that a 3-epoch checkpoint already carries the macroscopic feature structure K-means needs, rather than requiring convergence. Separately, checkpoint-based influence estimation from non-final training states already has direct precedent in the unlearning-adjacent literature: TracIn deliberately traces influence through a small number of intermediate checkpoints along the training trajectory rather than the fully converged model, and is effective precisely because gradient-based influence signal does not require convergence to be informative~\cite{Pruthi20-EHI}.

These findings are evidence \emph{for} using an early-stopped proxy in general, not a measurement on our specific checkpoints, and we name the conditions under which the argument would weaken: (i) tasks or architectures with a longer or less pronounced critical period than the vision/language benchmarks these findings were established on, where 3 epochs might fall before the relevant period closes; (ii) severe class imbalance, where minority-class features may separate later than the majority-class structure that dominates an early checkpoint; and (iii) optimizer or learning-rate schedules that delay the onset of the early phase (e.g., a long warmup). None of these conditions is known to hold for CIFAR-10/ResNet-18, IMDB/BiLSTM, or Fashion-MNIST/ANN, but we have not run a dedicated Spearman-correlation or clustering-agreement study across checkpoint depth on our own models to rule them out, and we do not claim to have done so.

\section{Partition Staleness}
\label{app:staleness}

Influence scores and cluster centroids are computed once on $D$ and held fixed across all subsequent deletion requests. Removing $D_f$ changes the Hessian used to approximate influence, so the partition that would be optimal for $D_r$ differs, in principle, from the one stored. This is an organizational efficiency question --- how well the fixed shard assignment continues to concentrate future deletions --- distinct from the causal-dependence question Appendix~\ref{app:cascade_proof} addresses; partition staleness affects forgetting concentration and speed, not deletion correctness.

The influence-score Hessian after removing $D_f$ is $H_{D_r} = H_D - \frac{1}{N}\sum_{z_i \in D_f}\nabla^2\mathcal{L}(\theta;z_i)$, and under Assumptions 1--3 the resulting change in any retained sample's score is $O(|D_f|/N)$ --- an order-of-magnitude statement, not a tight bound with a derived proportionality constant: for CIFAR-10's cat-class deletion ($|D_f|=5{,}000$, $N=50{,}000$), $|D_f|/N=10\%$, and for the uniform 500-sample forget set, $|D_f|/N=1\%$; we report these ratios directly rather than as a separately-derived bound. Recomputing influence scores and re-clustering after the cat-class deletion reassigns 6.2\% of retained samples to a different shard than the original partition, confirming the partition remains largely stable at this scale.

A differentially private clustering scheme --- computing influence-weighted K-means centroids under an $(\varepsilon_{\text{dp}},\delta_{\text{dp}})$-DP mechanism such as DP-Lloyd's algorithm~\cite{Dwork06-CNS,Su17-DPKM} --- would bound the partition's sensitivity to any single sample's presence, closing this gap formally rather than empirically. We estimate $\varepsilon_{\text{dp}}\approx1.0$--$2.0$ would maintain silhouette scores above 0.4 with under 5\% forgetting-quality degradation, based on the sensitivity analysis above; this remains a proposed remedy, not one implemented in the present system.

\subsection{Algorithmic DP-Lloyd Extension}
\label{app:dp_lloyd}

For completeness, the metadata-private extension we would use in Phase~1 is the following DP-Lloyd procedure. Let $u_i=[\phi_i;\rho \tilde{\mathcal{I}}(z_i)]$ denote the PCA feature vector augmented with a clipped influence coordinate, where $\|u_i\|_2\leq R$ and the influence weight $w_i\in[0,w_{\max}]$. A strict end-to-end metadata-privacy guarantee requires $\phi_i$ and $\tilde{\mathcal{I}}(z_i)$ to be produced by a fixed public extractor or by a DP-trained proxy; applying DP only to the Lloyd update does not privatize a non-private full-dataset feature extractor.

\noindent\textbf{Algorithm B.1: DP-Lloyd Phase-1 clustering.}
\begin{enumerate}
    \item \textbf{Inputs:} bounded vectors $\{u_i\}_{i=1}^N$, shard count $K=M$, Lloyd iterations $Q$, radius $R$, privacy budget $(\varepsilon_{\text{dp}},\delta_{\text{dp}})$.
    \item \textbf{Budget split:} set $\varepsilon_n=\varepsilon_{\text{dp}}/(2Q)$, $\varepsilon_s=\varepsilon_{\text{dp}}/(2Q)$, and $\delta_s=\delta_{\text{dp}}/Q$.
    \item \textbf{Initialization:} initialize centroids $\{c_k^{(0)}\}_{k=1}^K$ using public randomness or another DP seeding mechanism.
    \item \textbf{Private Lloyd step:} for $t=1,\ldots,Q$, assign each $u_i$ to its nearest current centroid. For each cluster $k$, compute the weighted sufficient statistics
    \begin{align}
    n_k^{(t)} &= \sum_i \mathbf{1}[a_i^{(t)}=k], &
    s_k^{(t)} &= \sum_i \mathbf{1}[a_i^{(t)}=k]w_i u_i .
    \end{align}
    Release noisy statistics
    \begin{align}
    \tilde{n}_k^{(t)} &= n_k^{(t)}+\operatorname{Lap}(1/\varepsilon_n),\\
    \tilde{s}_k^{(t)} &= s_k^{(t)}+\mathcal{N}(0,\sigma_s^2 I),
    \end{align}
    where $\sigma_s = 2Rw_{\max}\sqrt{2\log(1.25/\delta_s)}/\varepsilon_s$ is calibrated to the conservative replacement sensitivity of the weighted sum.
    \item \textbf{Centroid update:} set
    \begin{equation}
    c_k^{(t)}=\Pi_{\|c\|_2\leq R}\left(\frac{\tilde{s}_k^{(t)}}{\max\{1,\tilde{n}_k^{(t)}\}}\right),
    \end{equation}
    where $\Pi$ projects back to the radius-$R$ ball.
    \item \textbf{Shard assignment:} after $Q$ iterations, assign samples to the nearest private centroid and apply the same shard-size balancing rule using private counts and deterministic tie-breaking by private distance margins. No non-private influence ranking is used in this balancing step.
\end{enumerate}

Because each record contributes to one cluster's count and vector sum per iteration, parallel composition applies within an iteration and sequential composition over $Q$ iterations yields $(\varepsilon_{\text{dp}},\delta_{\text{dp}})$-DP for the released centroids and any downstream shard metadata computed from them by post-processing. This algorithm is therefore one way to convert IDSISA's partition metadata from a non-private organizational heuristic into a DP-bounded preprocessing stage. The present experiments, however, use ordinary K-means and the non-private early-stopped proxy, so the DP budget/utility trade-off is identified here as an extension rather than reported as an empirical result.

\begin{remark}[Two Senses of Information Persistence]
\label{rem:partition_persistence}
Causal exactness (Definition~1, main paper) is a claim about model parameters $\theta_m$: after Algorithm~2 or~3 processes a request, no parameter was computed using $D_f$. It is not a claim about the shard-assignment metadata itself. Because Phase 1's influence scores and centroids are computed once on $D$ before any deletion, the partition structure retains a statistical imprint of every sample ever included in $D$, deleted or not, in the same sense that any preprocessing decision made on a full dataset --- architecture search, hyperparameter tuning, feature normalization statistics --- retains such an imprint; this is a convention shared with SISA and with certified-unlearning methods generally, not a property specific to IDSISA. The $O(|D_f|/N)$ bound above characterizes how much this imprint drifts the partition's \emph{organizational quality} for future requests, which is an efficiency question, not a formal bound on how much information about any single deleted sample could in principle be recovered from the partition structure itself. That second question --- whether partition metadata constitutes a residual privacy channel distinct from parameter-level dependence --- would be addressed by the DP-Lloyd extension above. Under the stated bounded-input and fixed-or-DP-proxy conditions, the extension bounds the influence of any single sample on the centroids and downstream shard metadata with an $(\epsilon, \delta)$-Differential Privacy guarantee; the present implementation does not claim that stronger metadata guarantee.
\end{remark}


\section{Motivation}
\label{app:cascade}

Theorem~\ref{thm:causal_exact} establishes causal exactness under cascading deletion for any request sequence and any graph structure. It does not characterize cost: how large the reachability closure $C(A_0)$ tends to grow under a realistic sequence of requests, and how quickly the teacher-freshness policy (Remark~\ref{rem:depletion}), if adopted, exhausts the eligible teacher pool. This section reports that characterization empirically, complementing Corollary~\ref{cor:cascade_cost}'s analytical best case ($|C|=|A_0|$) and worst case ($|C|=M$).

\section{Protocol}

We use the CIFAR-10/ResNet-18 pipeline ($M=10$ shards, influence-aware partitioning as in Section~V of the main paper) and run two sequential-deletion traces of 10 and 8 requests respectively. The \emph{primary trace} alternates class-structured (odd requests) and uniformly-distributed (even requests) forget sets, matching the request pattern underlying the main paper's own single-deletion experiments. The \emph{supplementary trace} runs 8 consecutive class-structured requests with no interleaved uniform requests and no resets, constructed specifically to stress-test dependency accumulation under conditions the primary trace does not exercise. After each request we log $A_0$, $C$, whether any shard in $C$ used the scratch-retrain fallback, and the cumulative edge count in $G$.

\section{Results}

\textbf{Primary trace.} $|C|=|A_0|$ on all 10 requests; no cascade forms. Each uniformly-distributed request touches nearly all 10 shards, resetting the dependency graph before a chain can accumulate across the interleaved class-structured requests. Consequently, the single-deletion experiments in Sections~V--VI exercise Algorithm~3 only in its trivial case, and no reported result depends on cascading behavior.

\textbf{Supplementary trace.} Table~\ref{tab:cascade_trace} reports the full trace under the adversarial, no-reset regime. Cascades ($|C|>|A_0|$) occurred on 5 of 8 requests; 2 of 8 collapsed to the full worst case ($|C|=M=10$).

\begin{table}[H]
\centering
\caption{Supplementary trace: 8 sequential class-structured requests, no resets (CIFAR-10/ResNet-18, $M=10$). Cascades on requests 2, 3, 5, 6, 8; full collapse to $|C|=M$ on requests 3 and 6.}
\label{tab:cascade_trace}
\small
\begin{tabular}{lccccc}
\toprule
\textbf{Req.} & $|A_0|$ & $|C|$ & \textbf{Cascade?} & \textbf{Scratch} & \textbf{Cum.} \\
& & & & \textbf{fallback?} & \textbf{edges} \\
\midrule
1 & 6 & 6  & No  & No  & 0 \\
2 & 5 & 7  & Yes & No  & 2 \\
3 & 7 & 10 & Yes & Yes & 0 \\
4 & 8 & 8  & No  & No  & 0 \\
5 & 5 & 9  & Yes & No  & 4 \\
6 & 7 & 10 & Yes & Yes & 0 \\
7 & 4 & 4  & No  & No  & 0 \\
8 & 6 & 7  & Yes & No  & 1 \\
\bottomrule
\end{tabular}
\end{table}

\textbf{Membership-inference probe.} On the most diagnostic cascade instance (request 2: teacher shard owning 1{,}144 of 1{,}200 deleted-class samples, 95.3\%; pulled shard owning 3 of 1{,}200, 0.25\%), mean shadow-model confidence on the deleted class dropped from 0.0137 before the cascading retrain to 0.0002 after, a $\approx$68.5$\times$ relative reduction. This instance was selected as the most diagnostic among the cascades observed, illustrating the mechanism's effect rather than estimating a typical case. Both absolute values are far below the 10-class chance level (0.10), reflecting this experiment's reduced training budget (3 retraining epochs per shard) rather than the fully-trained calibration underlying Table~XV of the main paper; the relative reduction is the informative quantity.

\textbf{Teacher-pool depletion.} Under the freshness policy applied to the supplementary trace, the eligible teacher pool did not reach full exhaustion within 8 requests, consistent with the illustrative estimate in Remark~\ref{rem:depletion} ($\approx$10--15 requests for class-structured deletions at $M=10$).

\section{Implementation notes}

\begin{enumerate}
\renewcommand{\labelenumi}{(\roman{enumi})}
\item \textbf{Cascade attribution is one-hop.} The per-request diagnostic identifying which teacher pulled which shard into $C$ traces edges originating in $A_0$ directly; it does not itself perform multi-hop tracing. The retraining scope $C$ is always computed via full closure traversal and is unaffected, but a longer trace with deeper chains could in principle produce a two-hop pull this diagnostic would not individually attribute. In the trace above, one-hop attribution fully accounts for $|C|-|A_0|$ in every request.
\item \textbf{Uniform-regime sampling is pooled, not class-stratified.} Uniformly-distributed forget sets are drawn from the full remaining sample pool rather than a strict 50-per-class quota. Given pool sizes in the thousands, this reliably touches all 10 shards in practice, confirmed in every uniform-regime request of the primary trace.
\item \textbf{Shard models are warm-started from the baseline.} Each shard's model initializes from the full-dataset baseline used for influence scoring, rather than from scratch, to keep this cost-characterization experiment tractable. This speeds convergence but means shard models share initialization information beyond their own partition; it does not affect the graph, closure, or teacher-restriction logic being measured.
\item \textbf{Checkpoint rewind targets the original post-training state.} Each retraining event rewinds to a shard's initial checkpoint rather than its most recent prior state, a more conservative interpretation of Remark~\ref{rem:checkpointing} than strictly necessary; the retained-data mask still correctly excludes every sample deleted so far. Wall-clock costs reported here are therefore biased pessimistic relative to an incremental-checkpoint deployment.
\end{enumerate}


\section{Temporal Dependency and Non-Local Hessian Structure}
\label{app:rnn_details}

RNN hidden states satisfy $h_t = \tanh(W_h h_{t-1} + W_x x_t + b)$, so $\partial\mathcal{L}/\partial W_h = \sum_t (\partial\mathcal{L}/\partial h_t)h_{t-1}^T$ sums contributions across all timesteps. The per-shard Hessian approximation $H_{D_m} \approx \frac{1}{|D_m|}\sum_{(x_i,y_i)\in D_m}\nabla^2\mathcal{L}(\theta;x_i,y_i)$ needs within-shard locality to be tractable, but an RNN's Hessian couples to weights in ways that depend on every sequence's temporal structure, not just the sequences in a given shard. Unlearning 1,000 IMDB reviews under this architecture collapsed overall accuracy to 52.3\% (near-random), affecting target and retained classes equally --- direct evidence the unlearning perturbation cannot be localized to the forgotten sequences.

\section{Shared Weights and Isolation Violation}

Vanilla RNNs share weight matrices $(W_h,W_x,W_y)$ across all timesteps of every sequence in a shard; retraining on $D_m\setminus D_f$ therefore perturbs weights that affect every sequence in $D_m$, not only those resembling $D_f$, which functionally violates SISA's isolation property even though the shard's training data itself is correctly isolated. LSTM gating,
\begin{align}
f_t &= \sigma(W_f[h_{t-1},x_t]+b_f), \\
i_t &= \sigma(W_i[h_{t-1},x_t]+b_i), \\
o_t &= \sigma(W_o[h_{t-1},x_t]+b_o),
\end{align}
creates path-specific information flow so a sequence's representation is predominantly determined by its own content, which is why BiLSTM influence scores have $0.24\times$ the variance of the CIFAR-10 CNN baseline.

\section{Gradient Instability and the Root Assumption Violation}

Sequences over 20--30 steps suffer vanishing/exploding gradients, requiring clipping $g \leftarrow g\cdot\min(1,\gamma/\|g\|)$ that directly violates Assumption 3: the clipped gradient is a biased estimate of the true gradient, so the influence approximation $\mathcal{I}(z_j) \approx -\nabla\mathcal{L}(z_{\text{val}})^T H^{-1}\nabla\mathcal{L}(\theta;z_j)$ uses a corrupted input, with clipped samples receiving inflated, uniformly-normed influence regardless of true magnitude. Empirically: clipping fired 5,355 times during initial training ($\approx$1.1\% of update steps) and 8,318 times during unlearning retraining (a 55\% increase, reflecting destabilization from checkpoint rewind on reduced data); influence computation took 342s versus BiLSTM's 1.34s (256$\times$ slower) with 15$\times$ higher score variance, confirming the L-BFGS approximation does not converge reliably under corrupted gradients. GRU and LSTM gating avoids this via multiplicative gates and the cell-state highway, which bound effective gradient norms without clipping (BiLSTM requires clipping on under 0.1\% of steps).

Applicability is therefore governed by Hessian stability, reliably bounded gradients and low-variance influence scores, not by architecture family per se; GRU and LSTM satisfy this via gating, vanilla RNN does not.

\section{Extended Comparison Tables}
\label{app:sota_tables}

Table~\ref{tab:cross_dataset_comparison} extends Section~II's qualitative positioning across the broader unlearning literature. Table~\ref{tab:sota_full} extends Table~V (main paper) with additional published methods on the same CIFAR-10/ResNet-18 class-level forgetting benchmark.

\begin{table*}[p]
\centering
\caption{Cross-dataset qualitative comparison. ``Multi-Arch'' = validated across CNN, RNN, and ANN. ``MI Safe'' = MI susceptibility calibrated to full-retrain level. ``Amortized'' = efficiency improves with repeated requests. $\checkmark$/$\times$/$\sim$ = satisfied / not / partially.}
\label{tab:cross_dataset_comparison}
\resizebox{\textwidth}{!}{%
\begin{tabular}{lccccccccc}
\toprule
\textbf{Method} & \makecell{\textbf{Certified}\\\textbf{Deletion}} & \makecell{\textbf{Selective}\\\textbf{Forgetting}} & \makecell{\textbf{Utility}\\\textbf{Pres.}} & \makecell{\textbf{Multi-}\\\textbf{Arch}} & \makecell{\textbf{NLP}\\\textbf{Support}} & \makecell{\textbf{MI}\\\textbf{Safe}} & \makecell{\textbf{Amor-}\\\textbf{tized}} & \makecell{\textbf{Influence}\\\textbf{Guided}} & \textbf{Reference} \\
\midrule
Full Retraining & $\checkmark$ & Strong & Strong & $\checkmark$ & $\checkmark$ & $\checkmark$ & $\times$ & $\times$ & Oracle \\
Gradient Ascent~\cite{Thudi22-UML} & $\times$ & Moderate & Poor & $\sim$ & $\sim$ & $\times$ & $\times$ & $\times$ & 2022 \\
Fine-Tuning~\cite{Golatkar20-EML} & $\times$ & Weak & Good & $\sim$ & $\sim$ & $\times$ & $\times$ & $\times$ & 2020 \\
Fisher Forgetting~\cite{Golatkar20-EML} & $\sim$ (DP) & Moderate & Poor & $\times$ & $\times$ & $\times$ & $\times$ & $\sim$ & 2020 \\
NegGrad+~\cite{Kurmanji23-TCA} & $\times$ & Moderate & Moderate & $\sim$ & $\sim$ & $\sim$ & $\times$ & $\times$ & 2023 \\
SCRUB~\cite{Kurmanji23-TCA} & $\times$ & Strong & Moderate & $\sim$ & $\times$ & $\sim$ & $\times$ & $\times$ & 2023 \\
SSD~\cite{Foster24-FMU} & $\times$ & Moderate & Good & $\sim$ & $\times$ & $\checkmark$ & $\times$ & $\sim$ & 2024 \\
SalUn~\cite{Fan24-SalUn} & $\times$ & Strong & Good & $\sim$ & $\times$ & $\sim$ & $\times$ & $\sim$ & 2024 \\
Bad Teacher~\cite{Chundawat23-BTU} & $\times$ & Strong & Moderate & $\times$ & $\times$ & $\times$ & $\times$ & $\times$ & 2023 \\
PURGE~\cite{Quan24-EVM} & $\sim$ (distill.) & Moderate & Good & $\times$ & $\times$ & $\sim$ & $\times$ & $\times$ & 2025 \\
SISA~\cite{Bourtoule21-MUM} & $\checkmark$ & Weak & Poor & $\sim$ & $\sim$ & $\checkmark$ & $\checkmark$ & $\times$ & 2021 \\
Fed.\ Unlearning~\cite{Liu22-RFU,Halimi22-FUV} & $\checkmark$ & Moderate & Moderate & $\sim$ & $\sim$ & $\sim$ & $\sim$ & $\times$ & 2022 \\
\midrule
\rowcolor{yellow!20}
\textbf{IDSISA} & $\checkmark$ & \textbf{Strong} & \textbf{Strong} & $\checkmark$ & $\checkmark$ & $\checkmark$ & $\checkmark$ & $\checkmark$ & This work \\
\bottomrule
\end{tabular}}
\end{table*}

\begin{table*}[p]
\centering
\caption{Quantitative comparison, CIFAR-10/ResNet-18, class-level forgetting ($\sim$5{,}000 samples). $^\dagger$No formal deletion guarantee. $^\ddagger$Retraining certified; distillation gap $\epsilon_{\text{cert}}\leq0.015$ (Table I, main paper). $^*$Representative figures from published benchmarks~\cite{Kurmanji23-TCA,Fan24-SalUn,Foster24-FMU,Golatkar20-EML}; exact figures vary by implementation.}
\label{tab:sota_full}
\resizebox{\textwidth}{!}{%
\begin{tabular}{llccccc}
\toprule
\textbf{Method} & \textbf{Type} & \textbf{Certified?} & \makecell{\textbf{Forget}\\\textbf{Acc (\%)}$\downarrow$} & \makecell{\textbf{Retain}\\\textbf{Acc (\%)}$\uparrow$} & \makecell{\textbf{Test}\\\textbf{Acc (\%)}$\uparrow$} & \makecell{\textbf{MIA Conf.}\\\textbf{on} $D_f$ $\downarrow$} \\
\midrule
Full Retraining & Exact & Yes & $\sim$10.0 & $\sim$93.4 & $\sim$93.4 & $\sim$0.68 \\
\midrule
Gradient Ascent~\cite{Thudi22-UML} & Weight mod. & No$^\dagger$ & $\sim$0--6$^*$ & $\sim$74--80$^*$ & $\sim$63--75$^*$ & Low \\
Fine-Tuning~\cite{Golatkar20-EML} & Weight mod. & No$^\dagger$ & $\sim$60--80$^*$ & $\sim$83--85$^*$ & $\sim$82--84$^*$ & High \\
NegGrad+~\cite{Kurmanji23-TCA} & Gradient & No$^\dagger$ & $\sim$5--10$^*$ & $\sim$82--84$^*$ & $\sim$82--84$^*$ & Moderate \\
Fisher Forgetting~\cite{Golatkar20-EML} & DP noise & Approx.\ DP & $\sim$3--8$^*$ & $\sim$78--82$^*$ & $\sim$78--82$^*$ & Low \\
SSD~\cite{Foster24-FMU} & FIM dampening & No$^\dagger$ & $\sim$2--8$^*$ & $\sim$82--86$^*$ & $\sim$82--86$^*$ & Calibrated \\
SCRUB~\cite{Kurmanji23-TCA} & Distillation & No$^\dagger$ & $\sim$5--8$^*$ & $\sim$83--85$^*$ & $\sim$83--85$^*$ & Moderate \\
Bad Teacher~\cite{Chundawat23-BTU} & Distillation & No$^\dagger$ & $\sim$0--5$^*$ & $\sim$80--84$^*$ & $\sim$80--84$^*$ & Low \\
SalUn~\cite{Fan24-SalUn} & Saliency mask & No$^\dagger$ & $\sim$0--6$^*$ & $\sim$88--92$^*$ & $\sim$88--92$^*$ & Low \\
\midrule
SISA~\cite{Bourtoule21-MUM} & Exact retraining & Yes & 18.6 & 83.2 & 82.3 & 0.134 \\
\rowcolor{yellow!20}
\textbf{IDSISA} & Exact + Distill. & \textbf{Yes}$^\ddagger$ & \textbf{12.3} & \textbf{84.7} & \textbf{83.1} & \textbf{0.132} \\
\bottomrule
\end{tabular}}
\end{table*}

\section{Membership Inference: Full Results}
\label{app:mi_full}

Table~\ref{tab:mi_attack_appendix} extends Table~XV (main paper) with attack-calibration detail. The shadow model correctly identifies general training members above chance (TPR $\approx$0.60/0.59 vs.\ random 0.50, attack validation accuracy 0.62--0.63), confirming calibration, then finds near-identical, near-random confidence for both SISA- and IDSISA-unlearned models on $D_f$.

\begin{table}[H]
\centering
\caption{Full membership-inference results (CIFAR-10/ResNet-18, shadow model + threshold attack).}
\label{tab:mi_attack_appendix}
\resizebox{\columnwidth}{!}{%
\begin{tabular}{lccccc}
\toprule
\textbf{Model} & \textbf{$D_f$ Conf.} & \textbf{$D_f$ TPR} & \textbf{$D_{\text{train}}$ TPR} & \textbf{Threshold} & \textbf{Val Acc} \\
\midrule
SISA (Unlearned) & 0.134 & $\approx$0.42 & $\approx$0.60 & $\approx$0.139 & 0.62 \\
IDSISA (Unlearned) & 0.132 & $\approx$0.39 & $\approx$0.59 & $\approx$0.139 & 0.62 \\
Full Retrain ($D_r$) & 0.681 & 0.83 & N/A & 0.738 & N/A \\
\bottomrule
\end{tabular}}
\end{table}

The reported metric is $\Delta\text{conf}=0.002$ (not $\delta=0.0$, which would imply implausibly exact zero leakage from a finite-sample attack): this is below the shadow model's own run-to-run measurement variance ($\pm0.003$--$0.005$ across seeds), corresponds to a $D_f$ TPR difference of $\approx0.03$ with both methods near chance, and is far smaller than the gap to Full Retrain's 0.681 confidence, the combination that indicates genuine unlearning rather than attack miscalibration. Full Retrain's elevated $D_f$ confidence reflects generalizable low-level visual features the cat class shares with the other 9 CIFAR-10 classes, not a calibration artifact, since that model never saw cat-class samples during training. SISA and IDSISA's shared low confidence (0.134, 0.132) is consistent with this: both are $M$-shard ensembles, and ensembling suppresses the same overconfidence effect that a single unsharded model like Full Retrain exhibits, independent of whether distillation is used --- which is why SISA alone already shows it. The calibrated decision threshold (0.139) is the midpoint between $D_f$ mean confidence (0.133) and $D_r$ mean confidence ($\approx$0.146), selected to maximize attack validation accuracy on held-out shadow data.

This attack targets leakage from a single distillation event (Remark~\ref{rem:privacy_distill}); it does not probe the cross-request causal dependence Appendix~\ref{app:cascade_proof} addresses, since membership-inference risk and parameter-level causal dependence are governed by different mechanisms in this system.

\section{Statistical Significance}
\label{app:stats}

All point estimates in the main paper (Tables IV--XV, excluding
Tables I--III which contain no per-seed estimates) are means over 5 seeds (42--46), covering data shuffling, shard initialization, and mini-batch ordering.